\documentclass[a4paper,10pt]{elsarticle}
\usepackage{lineno,hyperref}
\usepackage{amsmath}
\usepackage{amsfonts}
\usepackage{amsthm}
\usepackage{amssymb}
\usepackage{pdflscape}
\usepackage{threeparttable}
\usepackage{mathrsfs}
\usepackage{bm}
\usepackage{geometry}
\usepackage{rotating}
\usepackage{pifont}
\usepackage{threeparttable}
\usepackage{color}
\usepackage{diagbox}
\usepackage{makecell}
\usepackage{bm}
\usepackage[graphicx]{realboxes}

\usepackage[dvips]{epsfig}
\usepackage{epsfig, epsf}

\usepackage{algorithm}
\usepackage{algorithmic}
\usepackage{multirow}

\usepackage{longtable}

\usepackage{booktabs}

\usepackage{subfig}

\usepackage{acronym}
\acrodef{AVF}{Additive Value Function}
\acrodef{MVF}{Marginal Value Function}
\acrodef{MCDA}{Multiple Criteria Decision Aiding}
\acrodef{ML}{Machine Learning}
\acrodef{ANN}{Artificial Neural Network}
\acrodef{AI}{Artificial Intelligence}
\acrodef{DL}{Deep Learning}
\acrodef{OWA}{Ordered Weighted Average}
\acrodef{NFS}{Net Flow Score}
\acrodef{CAI}{Class Acceptability Index}
\acrodef{APOI}{Assignment-based Pairwise Outranking Index}
\acrodef{KLR}{Kernel Logistic Regression}
\acrodef{ROR}{Robust Ordinal Regression}
\acrodef{SOR}{Stochastic Ordinal Regression}
\acrodef{SMAA}{Stochastic Multicriteria Acceptability Analysis}
\acrodef{ERA}{Extreme Ranking Analysis}
\acrodef{DM}{Decision Maker}
\acrodef{MAVT}{Multi-Attribute Value Theory}
\acrodef{MILP}{Mixed-Integer Linear Programming}
\acrodef{LP}{Linear Programming}
\acrodef{MCMC}{Markov Chain Monte Carlo}
\acrodef{SA}{Sensitivity Analysis}
\acrodef{PFA}{Post Factum Analysis}
\acrodef{DEA}{Data Envelopment Analysis}
\acrodef{MOO}{Multiple Objective Optimization}
\acrodef{HAR}{Hit-And-Run}
\acrodef{OR}{Operational Research}
\acrodef{DEA}{Data Envelopment Analysis}
\acrodef{DMU}{Decision Making Unit}
\modulolinenumbers[5]

\newdefinition{definition}{Definition}

\newproof{pf}{Proof}

\usepackage{setspace}
\makeatletter
\def\ps@pprintTitle{%
 \let\@oddhead\@empty
 \let\@evenhead\@empty
 \def\@oddfoot{}%
 \let\@evenfoot\@oddfoot}
\makeatother

\begin{document}
	
	\begin{frontmatter}
		
		\title{Selection of a representative sorting model in a preference disaggregation setting: a~review of existing procedures, new proposals, and experimental comparison}


		\author[put]{Micha{\l} W\'ojcik}
		\ead{96michal.wojcik@gmail.com}
		
		\author[put]{Mi{\l}osz Kadzi{\'n}ski\corref{cor}}
		\ead{milosz.kadzinski@cs.put.poznan.pl}

		\author[put]{Krzysztof Ciomek}
		\ead{k.ciomek@gmail.com}		

		\address[put]{Institute of Computing Science, Poznan University of Technology, Piotrowo 2, 60-965 Pozna{\'n}, Poland}
		\cortext[cor]{Corresponding author: Institute of Computing Science, Pozna\'{n} University of Technology, Piotrowo 2, 60-965 Pozna\'{n}, Poland. Tel. +48-61 665 3022.}

		\begin{abstract}
		We consider preference disaggregation in the context of multiple criteria sorting. The value function parameters and thresholds separating the classes are inferred from the Decision Maker's (DM's) assignment examples. Given the multiplicity of sorting models compatible with indirect preferences, selecting a single, representative one can be conducted differently. We review several procedures for this purpose, aiming to identify the most discriminant, average, central, benevolent, aggressive, parsimonious, or robust models. Also, we present three novel procedures that implement the robust assignment rule in practice. They exploit stochastic acceptabilities and maximize the support given to the resulting assignments by all feasible sorting models. The performance of sixteen procedures is verified on problem instances with different complexities. The results of an experimental study indicate the most efficient procedure in terms of classification accuracy, reproducing the DM's model, and delivering the most robust assignments. These include approaches identifying differently interpreted centers of the feasible polyhedron and robust methods introduced in this paper. Moreover, we discuss how the performance of all procedures is affected by different numbers of classes, criteria, characteristic points, and reference assignments. Finally, we illustrate the use of all approaches in a study concerning the assessment of the green performance of European cities.
        \end{abstract}
		
		
	\end{frontmatter}

\section{Introduction}
\label{sec:introduction}
\noindent In multiple criteria sorting problems, alternatives need to be assigned to preference-ordered classes~\cite{Zopounidis02}. Each of them is pre-defined and associated with a precise semantic, implying the same subsequent treatment of alternatives placed in a given category. The presence of multiple, potentially conflicting criteria makes such ordinal classification problems challenging. For this reason, the field of \ac{MCDA} offers a variety of methods that support the \acp{DM} in carrying forward the solution process (see, e.g., \cite{BELAHCENE2018, Meyer2019, SOYLU2011}). They are helpful in problem structuring, preference elicitation, construction and exploitation of the preference model, and explaining the recommended assignments~\cite{Alvarez2021}. In recent years, the approaches adopting a preference disaggregation perspective have been prevailing~\cite{Doumpos2018}. They construct a sorting model using a regression-like scheme based on the DM's decision examples. Such approaches facilitate the solution process by lowering the cognitive effort on the part of \acp{DM} and not requiring a specialized knowledge required when directly specifying values of decision model parameters.

The most popular preference disaggregation sorting method is UTADIS~\cite{Devaud80}. It accepts indirect preference information in the form of assignment examples, specifying the desired classification for a subset of reference alternatives~\cite{Jacquet01PD}. Such holistic statements are translated into compatible parameters of an additive value function and thresholds separating the classes on a scale of a comprehensive value~\cite{Zopounidis00}. UTADIS has been appreciated in the MCDA community for using an intuitive sorting procedure with highly interpretable alternatives' scores and class thresholds, while at the same time being free of statistic hypotheses and restrictions~\cite{Dimitras2002, Siskos05}. Also, it makes use of both qualitative and quantitative criteria, differentiates between inter- and intra-criteria attractiveness, and provides means for interaction with the DMs who might review the model by changing or enriching their preferences~\cite{Kadzinski18Nano}. Such appealing features have motivated the practical use of UTADIS for solving real-world decision problems concerning, e.g., business failure prediction~\cite{Zopounidis99}, credit risk assessment~\cite{Zopounidis2001fc}, supplier classification~\cite{Manshadi2015}, sorting activities in civil construction~\cite{Palha16RORUTADIS}, assessment of mutual funds~\cite{PENDARAKI2005}, energy analysis and policy making~\cite{Diakoulaki1999}, and adoption of green chemistry principles in nanotechnology~\cite{Kadzinski18Nano}.

The basic variant of UTADIS has been extended in numerous ways. In particular, it was generalized to an example-based procedure where the classes are delimited implicitly by decision examples rather than class thresholds~\cite{Greco10GMSDIS}. Furthermore, a sequential classification technique, called M.H.DIS, was introduced in~\cite{MHDIS} to consider the assignment of not yet classified alternatives to the most preferred class in a stepwise fashion. Moreover, UTADIS was advanced to a robustness analysis framework, where a multiplicity of compatible sorting models are exploited to verify the stability of classification. In \ac{ROR}, all such models are translated into the necessary and possible results using mathematical programming~\cite{Greco10GMSDIS, Koksalan09}. In turn, in \ac{SOR} -- the Monte Carlo simulations derive a large, representative set of such models whose results are summarized in the form of stochastic acceptabilities~\cite{Kadzinski13SORDIS}. Also, many works proposed dedicated techniques for dealing with inconsistency of assignment examples. They aim at restoring the consistency~\cite{Mousseau2006}, minimize a misclassification error~\cite{Zopounidis00}, use preference models compatible with different preferential reducts~\cite{Kadzinski18Nano}, or incorporate contingent, inter-related models that altogether reconstruct the holistic preference information~\cite{Kadzinski2020CORT}. In the same spirit, some optimization techniques were devised for handling large sets of assignment examples~\cite{LIU2020963, LIU20191071}.

Further methodological advancements have been devoted to supporting preference elicitation, enriching incorporated models, and addressing various structures and types of handled decision problems. In~\cite{Kadzinski21ALSORT}, one proposed active learning strategies that minimize the number of assignment examples needed for arriving at a~sufficiently robust recommendation. Moreover, \cite{Kadzinski14RORUTADIS} introduced a unified framework handling preference information in the form of assignment-based pairwise comparisons and constraints on the category sizes~\cite{MousseauSize, Ozpeynirci2018} along with assignment examples, whereas \cite{LIU2020963} accounted for valued desired classifications. An additive value function used in UTADIS was extended to admit interactions between criteria~\cite{Liu2021}, non-monotonicity~\cite{Kadzinski20Nano, Kadzinski21Nano, LIU20191071} or polynomial character~\cite{PUTADIS} of marginal value functions. Furthermore, \cite{Corrente17MCHPUTADIS} adapted the method to a~hierarchical structure of criteria, whereas \cite{Kadzinski21Nano, Ulucan13} considered a multi-decision classification problem with many interrelated decision attributes. Finally, dedicated group decision methods were devised for handling preferences of multiple \acp{DM} and either arriving at a~collective recommendation~\cite{LIU2015} or investigating the spaces of consensus and disagreement observed in the group~\cite{Kadzinski12GROUP}. 

A variety of real-world applications and methodological developments confirms the status of UTADIS as one of the essential methods in \ac{MCDA}. This paper deals with procedures for selecting a single instance of the threshold-based value-driven sorting model. Since the polyhedron of all functions and thresholds compatible with the stated indirect preference information can be quite large~\cite{Greco10GMSDIS, Koksalan09}, such a selection can be performed in different ways. Despite there exist robustness analysis methods that avoid such an arbitrary selection, the topic is worthwhile for a few reasons. The analysis of a single sorting model instance is straightforward even for non-experts in \ac{MCDA}, being less abstract than that of the entire set of infinitely many such compatible instances. Moreover, such a model can be displayed to the DMs, who can analyze the curvatures of marginal value functions, importances of criteria, dispersion of class thresholds, alternatives' comprehensive scores, and margins of safety in the recommended univocal assignments~\cite{Kadzinski11REPDIS}. In this way, it constitutes a synthetic and intuitive solution to the sorting problem, supporting the validity of the derived recommendation or motivating reactions from DMs. 

We contribute to the literature in a three-fold way. First, we review different concepts underlying the selection of a representative sorting model in the context of UTADIS. The primary idea consists of choosing the most discriminant model in terms of the differences between comprehensive values of reference alternatives assigned to different classes and/or marginal values associated with consecutive characteristic points of per-criterion functions~\cite{Beuthe01}. Furthermore, we discuss the concept of optimizing the comprehensive values attained by all alternatives or controlling the slope of marginal functions~\cite{Branke2015}. Another methodological stream is oriented toward identifying a central model with the proviso that the concept of centrality is interpreted in various ways~\cite{Bous10, Doumpos2014}. Moreover, we refer to the mean models obtained by averaging either the extreme models compatible with the DM's preferences~\cite{Jacquet82UTA} or a large sample of uniformly distributed ones~\cite{Kadzinski18Nano}. The last postulate builds on the outcomes of robustness analysis by making use of necessary, possible~\cite{Kadzinski11REPDIS} or stochastic results~\cite{Kadzinski13SORDIS} to define the targets that should be emphasized in the representative case.

Our second contribution consists of proposing novel procedures for selecting a single sorting model representative in the sense of robustness preoccupation. Specifically, we refer to the outcomes of Stochastic Ordinal Regression in the form of \acp{CAI} and \acp{APOI}~\cite{Kadzinski13SORDIS}. They quantify the shares of compatible sorting models, confirming a given alternative's assignment to a particular class or supporting one alternative being assigned to a class at least as good as another. The representative model emphasizes the most frequent classifications of all alternatives, or the most common assignment-based preference relations for all pairs of alternatives, or both these objectives at once. Similar to~\cite{Kadzinski11REPDIS}, we refer to the ``one for all, all for one" motto by representing all compatible sorting models, which contribute to the definition of a representative one. However, we build on more informative and detailed outcomes in the form of stochastic acceptabilities~\cite{Tervonen2009} rather than the possible and necessary assignments that need to be confirmed by at least one or all compatible models, respectively. We illustrate all procedures, including the existing and newly introduced ones, on a single decision problem to clarify their operational steps. 

The third and most crucial contribution consists of a thorough experimental evaluation of the sixteen discussed procedures. The problem of choosing the ``best" sorting model in the preference disaggregation methods is ill-defined. However, one can consider some objective criteria for the meaningful comparison of various procedures. In particular, we account for five measures that make sense in the context of both using incomplete preference information concerning a subset of reference alternatives and the multiplicity of sorting models compatible with the DM's assignments examples. They concern (i) the ability for reconstructing reference classification for all alternatives, (ii) the capability of restoring the preference model in terms of trade-off weights of criteria, alternatives' comprehensive values, and class thresholds, and (iii) the robustness of derived assignments in terms of the support they are given by all compatible models. The experiment involves problems with different numbers of classes, criteria, characteristic points of marginal functions, and reference alternatives assigned by the DM to each class. We discuss the average results attained over all considered settings and the trends observed with increasing model's complexity and availability of preference information. In this regard, our study can be seen as a significant extension of the experiments discussed in~\cite{Doumpos2014}, where only four procedures have been compared in a~similar context.

The paper's remainder is organized in the following. Section~\ref{sec:utadis} reminds UTADIS and its robust counterparts. In Section~\ref{sec:representative}, we discuss various procedures for selecting a representative sorting model. Section~\ref{sec:illustrate} illustrates their use on a didactic example. In Section~\ref{sec:experiment}, we present the results of an extensive experimental comparison of different approaches. The last section concludes the paper. 

\section{Reminder on UTADIS and robustness analysis}
\label{sec:utadis}
\noindent The following notation is used in the paper:
\begin{itemize}
	\item $A = \{a_{1}, a_{2}, \ldots, a_{i}, \ldots, a_{n}\}$ -- a finite set of $n$ alternatives; each of them is evaluated in terms of $m$ criteria;
	\item $A^{R} = \{a^{*}_{1}, a^{*}_{2}, \ldots, a^{*}_{r}\}$ -- a finite set of $r$ reference alternatives; $A^{R} \subseteq A$;
	\item $G = \{g_{1}, g_{2}, \ldots, g_{j}, \ldots, g_{m}\}$ -- a finite set of $m$ evaluation criteria, $g_{j}: A \rightarrow \mathbb{R}$ for all $j \in J = \{1, \ldots, m \} $;  without loss of generality, we assume that all of criteria in $G$ are of gain type;
	\item $X_{j} = \{g_{j}(a_{i}), a_{i} \in A\}$ -- a finite set of performances of all alternatives in $A$ on criterion $g_{j}$;
	\item $x^{1}_{j}, x^{2}_{j}, \ldots, x^{n_{j}(A)}_{j}$ -- the ordered values of $X_{j}$, $x^{k-1}_{j} < x^{k}_{j}, k = 2, \ldots, n_{j}(A)$, where $n_{j}(A) = |X_{j}|$ and $n_{j}(A) \leq n$; thus, $X = \prod^{m}_{j = 1}X_{j}$ is the performance space; note that $X_{j}$ can also be enriched with the extreme values of the performance scale that are not attained by any alternative;
	\item $C_{1}, \ldots, C_{p}$ -- $p$ pre-defined and preference-ordered classes so that $C_{l}$ is preferred to $C_{l-1}$ for $l = 2, \ldots, p$. 
\end{itemize}

\noindent To compute the desirability of each alternative $a\in A$, UTADIS~\cite{Devaud80} considers an \ac{AVF}~\cite{Keeney93}:
\begin{equation}
U(a) = \sum_{j=1}^{m}u_{j}(g_{j}(a)), \forall a \in A,
\end{equation}
where $u_j$, $j=1,\ldots,m$, are \acp{MVF} being piece-wise linear monotonic and defined by a~pre-defined number $\gamma_{j}$ of equally distributed characteristic points $\beta^{1}_{j}, \beta^{2}_{j}, \ldots, \beta^{\gamma_{j}}$, such that:
\begin{equation}
\beta^{s}_{j} = x^{1}_{j} + (x^{n_{j}(A)}_{j} - x^{1}_{j}) \frac{s - 1}{\gamma_{j} - 1}, j = 1, \ldots, m, s = 1, \ldots, \gamma_{j}.
\end{equation}
A comprehensive value is normalized in the $[0, 1]$ range by assuming that $u_{j}(\beta^{1}_{j}) = 0$, for $j=1,\ldots, m$, and $\sum_{j=1,\ldots,m} u_{j}(\beta^{\gamma_{j}}_{j}) = 1$. To enable control over the difference between marginal values assigned to the subsequent characteristic points, we consider $\rho_{j}(s-1, s), j = 1, \ldots, m, s = 2, \ldots, \gamma_{j}$, and $\rho$ variables defined as follows:
\begin{eqnarray}
\label{eqn:monotonicity_1}
u_{j}(\beta^{s}_{j}) - u_{j}(\beta^{s - 1}_{j}) \geq \rho_{j}(s - 1, s), \; j = 1, \ldots, m, s = 2, \ldots, \gamma_{j},\\
\label{eqn:monotonicity_2}
\rho_{j}(s - 1, s) \geq \rho, \; j = 1, \ldots, m, s = 2, \ldots, \gamma_{j}.
\end{eqnarray}
In the basic setting, $\rho$ is fixed to zero, which implies fulfilment of the weak monotonicity constraints. The marginal value for performance $x^{k}_{j}  \in [\beta^{s}_{j}, \beta^{s + 1}_{j}]$ can be computed using a~linear interpolation:
\begin{equation}
%\begin{split}
\label{eqn:surroundings_points}
u_{j}(x^{k}_{j}) = u_{j}(\beta^{s}_{j}) + (u_{j}(\beta^{s + 1}_{j}) -  u_{j}(\beta^{s}_{j}))\frac{x^{k}_{j} - \beta^{s}_{j}}{\beta^{s + 1}_{j} - \beta^{s}_{j}}, \; j = 1, \ldots, m, \; k = 1, \ldots, n_{j}(A).
%\end{split}
\end{equation}
UTADIS incorporates a threshold-based sorting procedure, where each class $C_l$ is delimited by the lower $t_{l - 1}$ and upper $t_{l}$ thresholds defined on a scale of a comprehensive value $U$. For simplicity, we do not consider the lower threshold of the least preferred class $C_1$ and the upper threshold of the most preferred class $C_p$, which could be arbitrarily fixed to $t_0 = 0$ and $t_p > 1$. Hence, to derive the assignment for alternative $a \in A$, $U(a)$ is compared with a vector of $p-1$ thresholds $t = [t_1, \ldots, t_{l-1}, t_l, \ldots, t_{p-1}]$ such that $t_{1} \geq \varepsilon$, $t_{l} - t_{l - 1} \geq  \varepsilon$, for $l = 2, \ldots, p - 1$, and $1 - t_{p-1} \geq \varepsilon$, where $\varepsilon$ is an arbitrarily small positive value. 

In UTADIS, the parameters of an assumed sorting model are inferred from the DM's indirect preference information. It consists of the desired class assignments for reference alternatives in $A^R$:
\begin{equation}
\label{eqn:preferences}
\forall_{a^{*} \in A^{R}} \; a^{*} \rightarrow C_{l}, \; l \in \{1, \ldots, p\}.
\end{equation}
The assignment examples are reproduced via preference disaggregation that ensures a comprehensive value of each reference alternative $a^{*} \in A^{R}$ is within the range $[t_{l-1}, t_l)$ delimited by the lower and upper thresholds corresponding to its desired class $C_l$, i.e.:
\begin{eqnarray}
\forall_{a^{*} \in A^{R}} \; a^{*} \rightarrow C_{l}, l \in \{1, \ldots, p-1\}            \implies t_{l} - U(a^{*}) \geq \delta + \varepsilon,\\
\forall_{a^{*} \in A^{R}} \;  a^{*} \rightarrow C_{l}, l \in \{2, \ldots, p\}            \implies U(a^{*}) - t_{l-1} \geq  \delta.
\end{eqnarray}
Variable $\delta$ allows for controlling the distances of alternatives' comprehensive values from the class limits. In the basic setting, it is set to zero. Overall, a set $\mathcal{U}^{R}$ of compatible \acp{AVF} and class thresholds is defined by the following set $E^{A^{R}}$ of linear constraints: 
\begin{align}
\left.
\begin{array}{ll}
\left.
\begin{array}{ll}
u_{j}(\beta^{1}_{j}) = 0, j = 1, \ldots, m, \\
\sum_{j=1}^{m} u_{j}(\beta^{\gamma_{j}}_{j}) = 1, \\
\end{array}
\right\} (E^{N})\\
u_{j}(\beta^{s}_{j}) - u_{j}(\beta^{s - 1}_{j}) \geq \rho_{j}(s - 1, s), \; j = 1, \ldots, m, s = 2, \ldots, \gamma_{j}, \\
\rho_{j}(s - 1, s) \geq \rho, \; j = 1,  \ldots, m, s = 2, \ldots, \gamma_{j}, \\
\left.
\begin{array}{ll}
t_{1} \geq \varepsilon, \\
t_{l} - t_{l - 1} \geq  \varepsilon, \; l = 2, \ldots, p - 1,\\
1 - t_{p-1} \geq \varepsilon, \\
\end{array}
\right\} (E^{T})\\
\forall_{a^{*} \in A^{R}} \; a^{*} \rightarrow C_{l}, l \in \{1, \ldots, p-1\}            \implies t_{l} - U(a^{*}) \geq \delta + \varepsilon, \\
\forall_{a^{*} \in A^{R}} \; a^{*} \rightarrow C_{l}, l \in \{2, \ldots, p\}            \implies U(a^{*}) - t_{l-1} \geq  \delta,
\end{array}
\right\} (E^{A^{R}})\label{eqn:constraints}
\end{align}
\noindent where $\rho$ and  $\delta$ are equal to zero, and $\varepsilon$ is a small positive constant. Please note that constraints ensuring monotonicity of the thresholds ($t_{l} - t_{l - 1} \geq  \varepsilon$) are redundant when the DM assigns at least one alternative to each class.

Since the DM's preference information is incomplete, when $E^{A^{R}}$ is feasible, $\mathcal{U}^{R}$ consists of infinitely many sorting models. To choose one of them, one needs to optimize an objective function. In Section~\ref{sec:representative}, we discuss sixteen procedures that differ mainly with respect to considering different objectives and/or incorporating additional variables and constraints. In what follows, we discuss the approaches for robustness analysis, whose results will be exploited by some of these procedures.

\ac{ROR} verifies the possibility or necessity of certain relationships based on a set of all compatible sorting models. This requires checking the consistency of the basic constraint set $E^{A^{R}}$ with additional constraints representing a~verified hypothesis. In what follows, we focus on a weak assignment-based preference relation $\succsim^{\rightarrow} : A \times A \rightarrow \{0, 1\}$, defined as follows:
\begin{equation}
\forall_{a, b \in A} \forall_{l_{a}, l_{b} \in \{1, \ldots, p\}} a \succsim^{\rightarrow} b \iff a \rightarrow C_{l_{a}}, b \rightarrow C_{l_{b}}, l_a \geq l_b.
\end{equation}
Its necessary counterparts holds if $\succsim^{\rightarrow}$ is confirmed for all compatible sorting models, i.e.:
\begin{equation}
\forall_{a, b \in A} a \succsim^{\rightarrow, N} b \iff \forall_{U \in \mathcal{U}^{R}} a \succsim^{\rightarrow} b. \\
\end{equation}
Note that if $\neg(a \succ^{\rightarrow, N} b)$, there exists at least one compatible sorting model in $\mathcal{U}^{R}$ that assigns $b$ to a more preferred class than $a$. Relations $\sim^{\rightarrow}$ and $\succ^{\rightarrow}$ along with their robust extensions $\sim^{\rightarrow, N}$ and $\succ^{\rightarrow, N}$ can be defined analogously by checking if one alternative is assigned to the same or more preferred class as another. The truth of these relations is verified using linear programming~\cite{Kadzinski11REPDIS, Kadzinski13SORDIS}.

In \ac{SOR}, $\mathcal{U}^{R}$ is exploited with the Monte Carlo simulations to derive a set $S \subseteq \mathcal{U}^{R}$ of uniformly distributed compatible sorting models. In practice, $S \subset \mathcal{U}^{R}$ and $|S| \ll |\mathcal{U}^{R}|$. The results obtained for these models are summarized in the form of four stochastic acceptabilities: Class Acceptability Indices ($CAI$s) and Assignment-based Pair-wise Winning ($APWI$), Outranking ($APOI$) and Equality ($APEI$) Indices. $CAI$s quantify the share of compatible sorting models assigning $a \in A$ to class $C_l$ and its approximation $CAI'$ is defined as follows, i.e.:
\begin{equation}
\forall_{a \in A} \forall_{l \in \{1, \ldots, p\}} \; CAI'(a, C_{l}) = \frac{|\{U \in S: a \rightarrow C_{l}\}|}{|S|}.
\end{equation}
Furthermore, $APWI: A \times A \rightarrow \mathbb{R} \in [0, 1]$ is defined as the share of all models in $\mathcal{U}^{R}$, which classify one alternative into a more preferred class than another alternative. Its approximation $APWI'$ is computed in the following way:
\begin{equation}
\forall_{a, b \in A} \; APWI'(a, b) = \frac{|\{U \in S: a \rightarrow C_{l_{a}}, b \rightarrow C_{l_{b}}, l_a > l_b\}|}{|S|}.
\end{equation}
The remaining pairwise indices, i.e., $APOI$s and $APEI$s, are defined analogously by referring to the shares of models confirming that one alternative is assigned to a class, respectively, at least as good or the same as another. In this paper, we sample from set $ \mathcal{U}^{R}$ using the \ac{HAR} algorithm implemented in~\cite{CiomekHAR}. 

\section{Procedures for selecting a representative sorting model}
\label{sec:representative}
\noindent In this section, we review different concepts underlying the selection of a representative sorting model in the context of UTADIS. Their most distinctive feature consists in optimizing a unique objective function subject to the constraint set $E^{A^{R}}$ that defines a set of all compatible value functions and class thresholds. Some procedures focus only on selecting a value function. In this case, the thresholds are set in equal distances between extreme comprehensive values of reference alternatives assigned to each class.

\subsection{The most discriminant models}
\noindent Let us start with the max-min formulations that seek the most discriminant model parameters. In the context of multiple criteria ranking, this idea was first implemented in UTAMP1~\cite{Beuthe01}. When it comes to sorting, \textbf{UTADISMP1}~\cite{Doumpos2014, Kadzinski11REPDIS} postulates maximizing the minimal difference between comprehensive values of reference alternatives and their respective class thresholds, i.e.:
\begin{equation}
\begin{split}
\mbox{Maximize } \delta, \mbox{ s.t. } E^{A^{R}}.
\end{split}
\end{equation}
In this way, the gap between all consecutive classes is maximized, yielding a model that is away from the boundaries of the polyhedron of all compatible sorting model~\cite{Bous10}. As a result, the DM's assignment examples are reproduced in a bold and robust way~\cite{Doumpos2014}.

Another procedure is motivated by the ranking method, called UTAMP2~\cite{Beuthe01}. Apart from optimizing $\delta$, i.e., the distances between the comprehensive values and class thresholds, it maximizes the difference between marginal values assigned to all pairs of consecutive characteristic points. The problem solved by \textbf{UTADISMP2} is the following:
\begin{equation}
\begin{split}
\mbox{Maximize } \delta + \rho, \mbox{ s.t. } E^{A^{R}},
\end{split}
\end{equation}
where $\rho \geq 0$. The method has similar features to UTADISMP1, while favoring steeper linear components of marginal value functions. This prevents weakly monotonic functions with level parts or even neglecting some criteria whose marginal functions take zero values for all performances. The $\rho$ component is considered alone in~\textbf{UTADISMP3}, highlighting the differences in values of marginal functions even more:
\begin{equation}
\begin{split}
\mbox{Maximize } \rho, \mbox{ s.t. } E^{A^{R}}.
\end{split}
\end{equation}

\subsection{Parsimonious decision model}
\label{subsec:MSCVF}
\noindent UTADISMP3 impacts the shape of marginal value functions by desiring the most discriminant ones. In turn, \cite{Branke2015}~postulated selecting as linear MVFs as possible, i.e., functions minimally deviating from the linearity. The model corresponding to this idea is called a Minimal Slope Change Value Function, in short, \textbf{MSCVF}. It can be obtained by solving the following \ac{LP} model:
\begin{center}
	$\mbox{Minimize } \phi$,
\end{center}
\begin{align}
s.t.
\left.
\begin{array}{ll}
\quad E^{A^{R}} \\
\left.
\begin{array}{ll}
\frac{u_{j}(\beta^{k}_{j}) - u_{j}(\beta^{k-1}_{j})}{\beta^{k}_{j} - \beta^{k-1}_{j}} - \frac{u_{j}(\beta^{k-1}_{j}) - u_{j}(\beta^{k-2}_{j})}{\beta^{k-1}_{j} - \beta^{k-2}_{j}} \leq \phi \\
\frac{u_{j}(\beta^{k-1}_{j}) - u_{j}(\beta^{k-2}_{j})}{\beta^{k-1}_{j} - \beta^{k-2}_{j}} - \frac{u_{j}(\beta^{k}_{j}) - u_{j}(\beta^{k-1}_{j})}{\beta^{k}_{j} - \beta^{k-1}_{j}} \leq \phi \\
\end{array}
\right\}  \mbox{for } j = 1, \ldots, m, k = 3, \ldots, \gamma^{j}
\end{array}
\right\} (E^{A^{R}}_{MSCVF})\label{eqn:constraints_mscvf}
\end{align}
Note that the use of MSCVF makes sense when at least three characteristic points are considered on a given criterion. The above idea can be interpreted in terms of favoring a parsimonious decision model consistent with the Occam razor principle. It says that ``entities must not be multiplied beyond necessity", which can be intuitively interpreted as: "the simplest explanation is most likely the correct one"~\cite{Greco11Pars}.

\subsection{Benevolent and aggressive models}
\noindent The problem of ambiguity in the definition of a model that should be used for conducting an actual analysis exists in many sub-fields of \ac{OR}. For example, in the cross-efficiency~\cite{Doyle94} considered in \ac{DEA}, one needs to select a weight vector for which a given \ac{DMU} attains its maximal efficiency, and that will be subsequently used to evaluate all remaining \acp{DMU}. Such a choice can be conducted in different ways, but two approaches, called aggressive and benevolent, are prevailing. They aim at selecting the weights that minimize or maximize the sum of efficiencies of other DMUs, respectively. Such weights correspond to the competitive and cooperative settings. A similar approach can be adapted to the context of \ac{MCDA}. In particular, \cite{Branke2015} proposed to derive a Maximal Sum of the Scores Value Function, in short, \textbf{MAX-SVF}, by maximizing a~sum of comprehensive values for all reference alternatives:
\begin{equation}
\begin{split}
\mbox{Maximize } \sum_{a^{*} \in A^{R}} U(a^{*}), \mbox{ s.t. } E^{A^{R}}.
\end{split}
\end{equation}
Alternatively, we can opt for the Minimal Sum of the Scores Value Function (\textbf{MIN-SVF}) that sheds a negative light on all reference alternatives considered jointly by minimizing a~sum of their comprehensive scores:
\begin{equation}
\begin{split}
\mbox{Minimize } \sum_{a^{*} \in A^{R}} U(a^{*}), \mbox{ s.t. } E^{A^{R}}.
\end{split}
\end{equation}

\subsection{Average models}
\noindent Another appealing idea consists of conducting a post-optimality analysis, deriving a set of representative sorting models, and averaging them to form an approximation of the polyhedron's centroid model~\cite{Doumpos2014}. It has been implemented in two different ways.

\textbf{UTADIS-JLS} was motivated by the system of $2m$ extreme solutions originally considered in the context of ranking problems~\cite{Jacquet82UTA}. Each of them is obtained by minimizing or maximizing the greatest value attained by MVF for each criterion, i.e.:
\begin{equation}
\begin{split}
\mbox{Maximize } \text{/} \mbox{ Minimize } u_{j}(\beta^{\gamma_{j}}_{j}), \mbox{ s.t. } E^{A^{R}}.
\end{split}
\end{equation}
Note that $u_{j}(\beta^{\gamma_{j}}_{j})$ can be interpreted as a weight or a trade-off constant of criterion $g_j$.

A~disadvantage of UTADIS-JLS consists of accounting only for the extreme models. In~\cite{Kadzinski18Nano} the concept of finding an ``average" model was generalized by considering a large sample $S = \{U^{1}, U^{2}, \ldots\, U^{|S|}\}$ of models considered in \ac{SOR}. The \textbf{CENTROID} procedure is not based on optimization. It derives an average of all samples that can be considered a stochastic approximation of the central solution. This applies to both characteristic points of MVFs and class threshold values:
\begin{equation}
\begin{split}
&t_l = \frac{1}{|S|} \sum_{i = 1}^{|S|} t^{i}_{l}, \; l = 1, \ldots, p-1, \\
&u_{j}(\beta_{j}^{s}) = \frac{1}{|S|} \sum_{i = 1}^{|S|} u^{i}_{j}(\beta_{j}^{s}), \; j = 1, \ldots, m, s = 1, \ldots, \gamma^{j}.
\end{split}
\end{equation}
Such average models are claimed to more robust and less vulnerable to changes in the DM's assignment examples~\cite{Doumpos2014}.

\subsection{Central models}
\noindent Opting for an average model can be seen as a particular implementation of selecting a central model. However, the concept of centrality can be interpreted in different ways, two of which -- denoted \textbf{CHEBYSHEV} and \textbf{ACUTADIS} -- are discussed in this subsection. 
The Chebyshev center of a polyhedron is a mid-point of the largest Euclidean ball that fits in a~polyhedron. A model corresponding to such a center was proposed in \cite{Doumpos2014}. To determine it, one needs to maximize variable $r$ that is inscribed in each monotonicity and assignment-based constraint:
\begin{center}
	\mbox{Maximize } $r$,
\end{center}
\begin{align}
s.t.
\left.
\begin{array}{ll}
E^{N}, \; E^{T}, \\
u_{j}(\beta^{s}_{j}) - u_{j}(\beta^{s - 1}_{j}) - r \geq 0, j = 1, \ldots, m, s = 2, \ldots, \gamma_{j}, \\
\forall_{a^{*}_i \in A^{R}} a^{*}_i \rightarrow C_{l}, l \in \{1, \ldots, p-1\}            \implies t_{l} - U(a^{*}_i) - b_{i} r \geq 0, \\
\forall_{a^{*}_{i} \in A^{R}} a^{*}_{i} \rightarrow C_{l}, l \in \{2, \ldots, p\}            \implies U(a^{*}_{i}) - t_{l-1} - c_{i} r \geq  0,
\end{array}
\right\} (E^{A^{R}}_{CC})\label{eqn:constraints_cc}
\end{align}
\noindent where $b_{i}$ and $c_{i}$ are the Euclidean norms of the decision variables' (except $r$) coefficients in constraint in which they occur~\cite{Doumpos2014}. Such a solution can be considered central because it is equally distant from all essential inequality constraints.

\textbf{ACUTADIS} postulates selecting an analytic center rather than the Chebyshev one. It was originally proposed for ranking problems and adjusted to the scope of sorting in~\cite{Kadzinski11REPDIS}. It corresponds to the model maximizing the logarithmic barrier function of the slacks ($d_{i^{-}}, d_{i^{+}}, d_{js}$) involved in the essential constraints of $E^{A^{R}}$~\cite{Doumpos2014}:
\begin{center}
		\mbox{Maximize } $\sum\limits_{a^{*}_{i} \in A^{R}} (\log d_{i^{-}} + \log d_{i^{+}}) + \sum\limits_{j=1}^{m} \sum\limits_{s = 2}^{\gamma_{j}} \log d_{js}$,
\end{center}
\begin{align}
s.t.
\left.
\begin{array}{ll}
E^{N}, \; E^{T}, \\
u_{j}(\beta^{s}_{j}) - u_{j}(\beta^{s - 1}_{j}) = d_{js}, \;  j = 1, \ldots, m, s = 2, \ldots, \gamma_{j}, \\
\forall_{a^{*}_{i} \in A^{R}} a^{*}_{i} \rightarrow C_{l}, l \in \{1, \ldots, p-1\} \implies t_{l} - U(a^{*}) - \delta = d_{i^{-}}, \\
\forall_{a^{*}_{i} \in A^{R}} a^{*}_{i} \rightarrow C_{l}, l \in \{2, \ldots, p\} \implies U(a^{*}_{i}) - t_{l-1} - \delta = d_{i^{+}}.
\end{array}
\right\} (E^{A^{R}}_{AC})\label{eqn:acutadis}
\end{align}
The above \ac{LP} model can be solved using the Newton's method~\cite{Bous10}, always leading to a unique solution. 

\subsection{Robust models based on exact outcomes}
\noindent The methods for robustness analysis were developed to exploit a set of all compatible models~\cite{Koksalan09, Greco10GMSDIS}. The derived outcomes reflect the stability of the sorting recommendation. However, their use for real-world decision aiding indicated that it is not easy for some users to comprehend such robust results and an abstract concept of infinitely many compatible models. This motivated the development of procedures for selecting a representative sorting model that can be exhibited to the \acp{DM}. The primary idea consisted of representing all compatible sorting models that contribute to the definition of a representative one. In this way, one does not lose the advantage of knowing all compatible ones while gaining a model instance that can be used to analyze the impact of different criteria, separation of decision classes, and robustness in the sense of distances of alternatives' values from class thresholds.

In~\cite{Kadzinski11REPDIS}, two objectives were defined to emphasize the robustness concerns. They are based on exact robust outcomes computed with mathematical programming. On the one hand, for all pairs of alternatives such that one of them is assigned to a class at least as good as another for all feasible models and for at least one of them -- it is assigned to a class strictly better, the difference between their comprehensive values should be maximized: 
\begin{center}
	\mbox{Maximize } $\omega$,
\end{center}
\begin{align}
s.t.
\left.
\begin{array}{ll}
\quad E^{A^{R}}, \\
\quad U(a) - U(b) \geq \omega &\forall_{a, b \in A } (a \succsim^{\rightarrow , N} b) \land \neg (b \succsim^{\rightarrow , N} a).  \\
\end{array}
\right\} (E^{A^{R}}_{R-iterative_{I}})\label{eqn:constraints_ror_interactive_1}
\end{align}
On the other hand, the value difference should be minimized for all pairs of alternatives necessarily assigned to the same class. This can be conducted while respecting the optimization of the previous target (i.e., setting $\omega = \omega^{*}$):
\begin{center}
	\mbox{Minimize } $\lambda$,
\end{center}
\begin{align}
s.t.
\left.
\begin{array}{ll}
E^{A^{R}}_{R-iterative_{I}}, \\
\omega = \omega^{*}, \\
U(c) - U(d) \leq \lambda &\forall_{c, d \in A} (c \sim^{\rightarrow , N} d), \\
U(d) - U(c) \leq \lambda &\forall_{c, d \in A} (c \sim^{\rightarrow , N} d).
\end{array}
\right\} (E^{A^{R}}_{R-iterative_{II}})\label{eqn:constraints_ror_interactive_2}
\end{align}
The procedure which attains the two targets iteratively is denoted by \textbf{ROBUST-ITER}. An alternative approach, called \textbf{ROBUST-COMP}, accounts for these objectives at the same time by solving the following \ac{LP} model:
\begin{center}
	\mbox{Maximize } $\iota$,
\end{center}
\begin{align}
s.t.
\left.
\begin{array}{ll}
\quad E^{A^{R}}, \\
\left.
\begin{array}{ll}
U(a) - U(b) - \iota \geq U(c) - U(d), \\
U(a) - U(b) - \iota \geq U(d) - U(c), \\
\end{array}
\right\} \forall_{a, b, c, d \in A } (a \succsim^{\rightarrow , N} b) \land \neg (b \succsim^{\rightarrow , N} a) \land (c \sim^{\rightarrow , N} d). \\
\end{array}
\right\} (E^{A^{R}}_{R-compromise})\label{eqn:constraints_ror_compromise}
\end{align}

\subsection{Robust models based on stochastic outcomes}
\noindent A sorting model that is representative in terms of the robustness preoccupation can also be selected based on the stochastic outcomes computed by \ac{SOR}. The idea implemented in \textbf{REPDIS} consists of emphasizing the advantage of these alternatives, which are assigned to a better class than others for a greater share of compatible sorting models, i.e., $APWI'(a, b) > APWI'(b, a)$. This can be attained by maximizing the minimal value difference for pairs of alternatives satisfying the above condition:
\begin{center}
	\mbox{Maximize } $\omega$,
\end{center}
\begin{align}
s.t.
\left.
\begin{array}{ll}
\quad E^{A^{R}}, \\
\quad U(a) - U(b) \geq \omega(a, b), &\forall_{a, b \in A} APWI'(a, b) > APWI'(b, a), \\
\quad \omega(a, b) \geq \omega, &\forall_{a, b \in A} APWI'(a, b) > APWI'(b, a).
\end{array}
\right\} (E^{A^{R}}_{APWI'_{I}})\label{eqn:constraints_apwi_1}
\end{align}
In the second stage, one can optimize the sum of elementary value differences $\omega(a, b)$, while respecting the results of the first stage by setting $\omega = \omega^{*}$, i.e.:
\begin{center}
	\mbox{Maximize } $\sum_{\forall_{a, b \in A} APWI'(a, b) > APWI'(b, a)} \omega(a, b) \mbox{ s.t. } E^{A^{R}}_{APWI'_{I}} \cup \omega = \omega^{*}.$
\end{center}

In what follows, we discuss three novel approaches that exploit the stochastic acceptability indices for selecting a single, robust sorting model. The first method, called \textbf{CAI}, aims at maximizing the $CAI(a_i, C_l)'$ corresponding to the class assignment $C_{l}$ suggested for each alternative $a \in A$ by a given sorting model $U \in \mathcal{U}^{R}$, denoted by $a \rightarrow^{U} C_{l}$. Due to the intrinsic nature of $CAI$s, maximization involves the product of values for individual alternatives instead of a sum. The main reason is that the relationships between $CAI$s should be compared in terms of a ratio rather than a difference. For example, $CAI'(a, C_{1}) = 0.25$ and $CAI'(a, C_{2}) = 0.75$ indicate that $a \rightarrow C_{2}$ occurred three times more often than $a \rightarrow C_{1}$ in the space of compatible sorting models. The objective function can be formulated as follows:
\begin{center}
	$U^{*} = {\arg \max}_{U \in \mathcal{U}^{R}} \prod_{\forall_{a_{i} \in A} : a_{i} \rightarrow^{U} C_{l}} CAI'(a_{i}, C_{l}) $.
\end{center}
We will replace the above non-linear form with its linear counterpart. Specifically, we replace the product of numbers by the sum of their logarithms (note that $CAI'$ values are computed beforehand). The objective function needs to build on $CAI$s that correspond to the class assignments of alternatives suggested by the selected model. This is ensured by introducing binary variables $x_{il}$ that should be equal to one when $a_i \rightarrow^{U} C_l$ is satisfied. After these transformations, the following problem is obtained:
\begin{center}
	\mbox{Maximize } $\kappa_{cai}  =  \sum\limits_{a_{i} \in A} \sum\limits_{l=1}^{p} x_{il} \log (CAI'(a_{i}, C_{l}) + \mu)$,
\end{center}
\begin{align}
s.t.
\left.
\begin{array}{ll}
E^{A^{R}}, \\
\forall_{a_{i} \in A} \forall_{l \in \{2, \ldots, p\}} \quad U(a_{i}) - t_{l-1} - \delta - M x_{il} \geq -M, \\
\forall_{a_{i} \in A} \forall_{l \in \{1, \ldots, p-1\}} \quad t_l - U(a_{i}) - \delta - M x_{il} \geq \varepsilon -M, \\
\forall_{a_{i} \in A} \sum\limits_{l = 1}^{p} x_{il} = 1,\\
\forall_{a_{i} \in A} \forall_{l \in \{1, \ldots, p\}}  x_{il} \in \{0,1\},
\end{array}
\right\} (E^{A^{R}}_{CAI_{I}})\label{eqn:constraints_cai_1}
\end{align}
\noindent where $M >> 1$ is a large constant and $\mu$ an arbitrarily small value. The latter is needed to ensure that the method works correctly when some $CAI'$ is zero. Note that for any inequality in the form: $X - Mb \geq -M$, where $X$ is an expression whose value can be determined and $b$ is a~binary variable, $b$ may be equal to one only if $X \geq 0$. Hence, variable $x_{il}$ will be equal to one when the conditions justifying $a_{i} \rightarrow^{U} C_{l}$, i.e., $U(a_{i}) \geq t_{l-1}$ and $t_l > U(a_{i})$, are met. Then, other variables $x_{ih}$, $h\ne l$, will be set to zero, hence satisfying the following constraint $\sum\limits_{l = 1}^{p} x_{il} = 1$.

Solving the above \ac{LP} problem allows identifying a sorting model that best represents the entire space in terms of assignments of alternatives to classes, measured with $CAI$s. As a secondary objective, we will regularize the model to balance the maximal shares of all criteria in the comprehensive value, hence advocating for a more central function. Specifically, we will minimize the deviations between the greatest marginal values for all pairs of criteria: 
\begin{center}
	\mbox{Minimize } $\xi$,
\end{center}
\begin{align}
s.t.
\left.
\begin{array}{ll}
E^{A^{R}}_{CAI_{I}}, \\
\kappa_{cai} = \kappa_{cai}^{*}, \\
\forall_{i, j \in \{1, \ldots, m\} \land i \neq j} \quad u_{i}(\beta^{\gamma_{i}}_{i}) - u_{j}(\beta^{\gamma_{j}}_{j}) \leq \xi, \\
\forall_{i, j \in \{1, \ldots, m\} \land i \neq j} \quad u_{j}(\beta^{\gamma_{j}}_{j}) - u_{i}(\beta^{\gamma_{i}}_{i}) \leq \xi.
\end{array}
\right\} (E^{A^{R}}_{CAI_{II}})\label{eqn:constraints_cai_2}
\end{align}
This secondary target will also be considered in the context of the following two procedures. Since the model used for this purpose will be the same, we will not repeat it to save space.

An analogous approach, called \textbf{APOI}, can be formulated based on the analysis of the stability of assignment-based relations for all pairs of alternatives rather than class assignments of individual alternatives. In particular, we will consider the following stochastic acceptabilities for all pairs of alternatives $(a,b) \in A\times A$:
\begin{itemize}
	\item $APWI'(a, b)$ indicating the share of models for which $a$ is assigned to a more preferred class than $b$, i.e., $l_a > l_b$;
	\item $APEI'(a, b)$ indicating the share of models for which $a$ is assigned to the same class as $b$, i.e., $l_a = l_b$;
	\item $APWI'(b,a)$ indicating the share of models for which $a$ is assigned to a less preferred class than $b$, i.e., $l_a < l_b$.
\end{itemize}
Overall, we aim at identifying the model emphasizing the assignment-based pairwise relations captured with $APWI'$s and $APEI'$s in the best way, i.e.:
\begin{center}
	$U^{*} = {\arg \max}_{U \in \mathcal{U}^{R}} \prod_{(a_i, a_j) \in A \times A: i \neq j}
	\begin{cases}
	APWI'(a_i, a_j) & \text{if } a_{i} \rightarrow^{U} C_{l_i} \land a_{j} \rightarrow^{U} C_{l_j} \land l_i > l_j,\\
	APEI'(a_i, a_j) & \text{if } a_{i} \rightarrow^{U} C_{l_i} \land a_{j} \rightarrow^{U} C_{l_j} \land l_i = l_j,\\
	APWI'(a_j, a_i) & \text{if } a_{i} \rightarrow^{U} C_{l_i} \land a_{j} \rightarrow^{U} C_{l_j} \land l_i < l_j.
	\end{cases}$
\end{center}

\noindent Similar to the CAI procedure, we introduce the binary variables corresponding to the three possible relations for each pair of alternatives $(a_{i}, a_{j}) \in A  \times A$, $i \neq j$: $v_{ij}$ corresponding to a scenario with $a_i$ being assigned to a more preferred class than $a_j$ (for the inverse situation, we consider $v_{ji}$) and $e_{ij}$ standing for $a_i$ and $a_j$ being assigned to the same class. After transforming the product of elementary objectives into the sum of respective logarithms, the folowing \ac{LP} model can be formulated:
\begin{equation*}
\begin{split}
	\mbox{Maximize } \kappa_{apoi} = 
\sum\limits_{(a_{i}, a_{j}) \in A \times A: i \neq j} v_{ij} \log (APWI'(a_{i}, a_{j}) + \mu)
+ \sum\limits_{(a_{i}, a_{j}) \in A \times A: i \neq j} e_{ij} \log (APEI'(a_{i}, a_{j}) + \mu) \\
+ \sum\limits_{(a_{i}, a_{j}) \in A \times A: i \neq j} v_{ji} \log (APWI'(a_{j}, a_{i}) + \mu),
\end{split}
\end{equation*}
\begin{align}
s.t.
\left.
\begin{array}{ll}
E^{A^{R}}_{CAI} \\
\forall_{(a_{i}, a_{j}) \in A  \times A: i \neq j} \sum\limits_{l=1}^{p} l x_{il} - \sum\limits_{l=1}^{p} l x_{jl} - M v_{ij} \geq 0.5 - M, \\
\forall_{(a_{i}, a_{j}) \in A  \times A: i \neq j} \sum\limits_{l=1}^{p} l x_{il} - \sum\limits_{l=1}^{p} l x_{jl} - M v_{ij} \leq 0.5, \\
\forall_{(a_{i}, a_{j}) \in A \times A: i \neq j} v_{ij} + e_{ij} + v_{ji} = 1,\\
\forall_{(a_{i}, a_{j}) \in A \times A: i \neq j} v_{ij}, e_{ij}, v_{ji} \in \{0,1\}.
\end{array}
\right\} (E^{A^{R}}_{APOI_{I}})\label{eqn:constraints_apoi_1}
\end{align}
The roles of $\mu$ and $M$ are the same as in the CAI procedure. The first of the three constraints included above enforces $v_{ij} = 0$ when $a_i$ is not assigned a class better than $a_j$. However, if $a_i$ is assigned to a more preferred class than $a_j$, then the second constraint enforces $v_{ij} = 1$. In case both $v_{ij} = 0$ and $v_{ji} = 0$, the third constraint would imply $e_{ij} = 1$. The three variables are used to select the factor in the maximization function for each pair of alternatives. In this way, the optimization focuses on assigning alternatives to classes to reflect as closely as possible the relationships between pairs of alternatives in the entire set of sorting models compatible with DM's preferences.

The joint focus on reproducing the most frequent assignments of individual alternatives and the most supported assignment-based preference relations is reflected in the \textbf{COMB} procedure. It combines the objective functions considered in \textbf{CAI} and \textbf{APOI} under a unified framework, hence reconciling between the two perspectives:
\begin{equation}
\begin{split}
	\mbox{Maximize } \; \kappa_{comb} =
\sum\limits_{a_{i} \in A} \sum\limits_{l=1}^{p} x_{il} \log (CAI'(a_{i}, C_{l}) + \mu) 
+ \sum\limits_{(a_{i}, a_{j}) \in A: i \neq j} v_{ij} \log (APWI'(a_{i}, a_{j}) + \mu) \\
+ \sum\limits_{(a_{i}, a_{j}) \in A: i \neq j} e_{ij} \log (APEI'(a_{i}, a_{j}) + \mu)
+ \sum\limits_{(a_{i}, a_{j}) \in A: i \neq j} v_{ji} \log (APWI'(a_{j}, a_{i}) + \mu), \mbox{   s.t. } E^{A^{R}}_{APOI_{I}}.
\end{split}
\end{equation}
%\begin{align}
%s.t. 
%\left.
%\begin{array}{ll}
%\quad E^{A^{R}}_{APOI_{I}}
%\end{array}
%\right\} (E^{A^{R}}_{COMB_{I}})\label{eqn:constraints_cai_apoi_1}
%\end{align}
Still, the idea of reflecting the outcomes of \ac{SOR} in a single model that can be exhibited to the \ac{DM} is maintained. 

\section{Illustrative study}
\label{sec:illustrate}
\noindent To illustrate how the procedures for selecting a representative sorting model work, we consider an example problem concerning the evaluation of $30$ major European cities in implementing green policy~\cite{EIU09}. Each city is rated in terms of the following four criteria: $CO_{2}$ emissions ($g_{1}$), energy consumption ($g_{2}$), water management ($g_{3}$), and waste and land use ($g_{4}$). The performances on the scale between $0$ and $10$ were determined by considering various indicators. They are given in Table~\ref{tab:case_study_dataset}. We will employ UTADIS with the aim of assigning the cities to three classes: $C_1$, $C_2$, and $C_3$, where $C_3$ is the most preferred category. We assume that a marginal function for each criterion has three characteristic points ($\gamma_{j} = 3$ for $j=1,\ldots,4$). Moreover, they are defined over the $[0,10]$ range, and thus $\beta^{1}_{j} = 0$, $\beta^{2}_{j} = 5$, and $\beta^{3}_{j} = 10$. 
\begin{table}[h]
	\caption{Evaluation of decision alternatives (cities) on four criteria, their marginal and comprehensive values according to a reference model, and Class Acceptability Indices $CAI'(a_{i}, C_{l})$ for all alternatives and classes.}\label{tab:case_study_dataset}\centering
	\footnotesize
	\begin{tabular} {|l||l|l|l|l||l|l|l|l||l||l|l|l|}
		\hline
		& \multicolumn{4}{c||}{Performances} & \multicolumn{5}{c||}{Reference values} & \multicolumn{3}{c|}{Class acceptabilities}\\\hline
		\textbf{Alternative} & \boldsymbol{$g_{1}$} & \boldsymbol{$g_{2}$} & \boldsymbol{$g_{3}$} & \boldsymbol{$g_{4}$} & \boldsymbol{$u_{1}$} & \boldsymbol{$u_{2}$} & \boldsymbol{$u_{3}$} & \boldsymbol{$u_{4}$} & \boldsymbol{$U(a)$} & \boldsymbol{$C_{1}$} & \boldsymbol{$C_{2}$} & \boldsymbol{$C_{3}$} \\ \hline
		$a_{1}$ (Oslo) & 9.58 & 8.71 & 6.85 & 8.23 & 0.0691 & 0.2384 & 0.1896 & 0.1992 & 0.6963  & 0.000 & 0.000 & 1.000 \\ \hline
		$a_{2}$ (Stockholm) & 8.99 & 7.61 & 7.14 & 7.99  & 0.0616 & 0.2240 & 0.2074 & 0.1872 & 0.6802 & 0.000 & 0.029 & 0.971 \\ \hline
		$a_{3}$ (Zurich) & 8.48 & 6.92 & 8.88 & 8.82 & 0.0552 & 0.2149 & 0.3142 & 0.2286 & 0.8129  & 0.000 & 0.000 & 1.000 \\ \hline
		$a_{4}$ (Copenhagen) & 8.35 & 8.69 & 8.88 & 8.05 & 0.0535 & 0.2381 & 0.3142 & 0.1902 & 0.7960 & 0.000 & 0.000 & 1.000 \\ \hline
		$a_{5}$ (Brussels) & 8.32 & 6.19 & 9.05 & 7.26 & 0.0532 & 0.2054 & 0.3246 & 0.1508 & 0.7339 & 0.000 & 0.000 & 1.000 \\ \hline
		$a_{6}$ (Paris) & 7.81 & 4.66 & 8.55 & 6.72 & 0.0467 & 0.1769 & 0.2939 & 0.1239 & 0.6414 & 0.000 & 0.056 & 0.944 \\ \hline
		$a_{7}$ (Rome) & 7.57 & 6.40 & 6.88 & 5.96 & 0.0437 & 0.2081 & 0.1915 & 0.0859 & 0.5292 & 0.000 & 1.000 & 0.000 \\ \hline
		$a_{8}$ (Vienna) & 7.53 & 7.76 & 9.13 & 8.60 & 0.0432 & 0.2259 & 0.3295 & 0.2177 & 0.8163 & 0.000 & 0.000 & 1.000 \\ \hline
		$a_{9}$ (Madrid) & 7.51 & 5.52 & 8.59 & 5.85  & 0.0429 & 0.1966 & 0.2964 & 0.0804 & 0.6163 & 0.000 & 0.124 & 0.876 \\ \hline
		$a_{10}$ (London) & 7.34 & 5.64 & 8.58 & 7.16 & 0.0408 & 0.1982 & 0.2958 & 0.1458 & 0.6805 & 0.000 & 0.000 & 1.000 \\ \hline
		$a_{11}$ (Helsinki) & 7.30 & 4.49 & 7.92 & 8.69 & 0.0403 & 0.1704 & 0.2553 & 0.2222 & 0.6881 & 0.000 & 0.200 & 0.800 \\ \hline
		$a_{12}$ (Amsterdam) & 7.10 & 7.08 & 9.21 & 8.98 & 0.0378 & 0.2170 & 0.3344 & 0.2366 & 0.8258 & 0.000 & 0.000 & 1.000 \\ \hline
		$a_{13}$ (Berlin) & 6.75 & 5.48 & 9.12 & 8.63 & 0.0334 & 0.1961 & 0.3289 & 0.2192 & 0.7775 & 0.000 & 0.000 & 1.000 \\ \hline
		$a_{14}$ (Ljubljana) & 6.67 & 2.23 & 4.19 & 5.95 & 0.0323 & 0.0846 & 0.0638 & 0.0854 & 0.2662 & 1.000 & 0.000 & 0.000 \\ \hline
		$a_{15}$ (Riga) & 5.55 & 3.53 & 6.43 & 5.72 & 0.0182 & 0.1340 & 0.1639 & 0.0739 & 0.3900 & 1.000 & 0.000 & 0.000 \\ \hline
		$a_{16}$ (Istanbul) & 4.86 & 5.55 & 5.59 & 4.86& 0.0110 & 0.1970 & 0.1124 & 0.0370 & 0.3573 & 1.000 & 0.000 & 0.000 \\ \hline
		$a_{17}$ (Athens) & 4.85 & 4.94 & 7.26 & 5.33 & 0.0109 & 0.1875 & 0.2148 & 0.0545 & 0.4677 & 0.088 & 0.912 & 0.000 \\ \hline
		$a_{18}$ (Budapest) & 4.85 & 2.43 & 6.97 & 6.27 & 0.0109 & 0.0922 & 0.1970 & 0.1014 & 0.4016 & 0.000 & 1.000 & 0.000 \\ \hline
		$a_{19}$ (Dublin) & 4.77 & 4.55 & 7.14 & 6.38 & 0.0108 & 0.1727 & 0.2074 & 0.1069 & 0.4978 & 0.000 & 1.000 & 0.000 \\ \hline
		$a_{20}$ (Warsaw) & 4.65 & 5.29 & 4.90 & 5.17 & 0.0105 & 0.1936 & 0.0747 & 0.0465 & 0.3252& 1.000 & 0.000 & 0.000 \\ \hline
		$a_{21}$ (Bratislava) & 4.54 & 4.19 & 7.65 & 5.60 & 0.0102 & 0.1590 & 0.2387 & 0.0680 & 0.4759 & 0.001 & 0.999 & 0.000 \\ \hline
		$a_{22}$ (Lisbon) & 4.05 & 5.77 & 5.42 & 5.34& 0.0091 & 0.1999 & 0.1019 & 0.0550 & 0.3659 & 1.000 & 0.000 & 0.000 \\ \hline
		$a_{23}$ (Vilnius) & 3.91 & 2.39 & 7.71 & 7.31 & 0.0088 & 0.0907 & 0.2424 & 0.1533 & 0.4952 & 0.000 & 0.935 & 0.065 \\ \hline
		$a_{24}$ (Bucharest) & 3.65 & 3.42 & 4.07 & 3.62  & 0.0082 & 0.1298 & 0.0620 & 0.0275 & 0.2276 & 1.000 & 0.000 & 0.000 \\ \hline
		$a_{25}$ (Prague) & 3.44 & 3.26 & 8.39 & 6.30  & 0.0078 & 0.1237 & 0.2841 & 0.1029 & 0.5185  & 0.000 & 0.849 & 0.151 \\ \hline
		$a_{26}$ (Tallinn) & 3.40 & 1.70 & 7.90 & 6.15 & 0.0077 & 0.0645 & 0.2541 & 0.0954 & 0.4216 & 0.008 & 0.991 & 0.001 \\ \hline
		$a_{27}$ (Zagreb) & 3.20 & 4.34 & 4.43 & 4.04  & 0.0072 & 0.1647 & 0.0675 & 0.0307 & 0.2701 & 1.000 & 0.000 & 0.000 \\ \hline
		$a_{28}$ (Belgrade) & 3.15 & 4.65 & 3.90 & 4.30 & 0.0071 & 0.1765 & 0.0594 & 0.0327 & 0.2757  & 1.000 & 0.000 & 0.000 \\ \hline
		$a_{29}$ (Sofia) & 2.95 & 2.16 & 1.83 & 3.32 & 0.0067 & 0.0820 & 0.0279 & 0.0252 & 0.1418& 1.000 & 0.000 & 0.000 \\ \hline
		$a_{30}$ (Kiev) & 2.49 & 1.50 & 5.96 & 1.43  & 0.0056 & 0.0569 & 0.1351 & 0.0109 & 0.2085 & 1.000 & 0.000 & 0.000 \\ \hline
	\end{tabular}
\end{table}

\begin{figure}[t]
\centering\includegraphics[width=10cm]{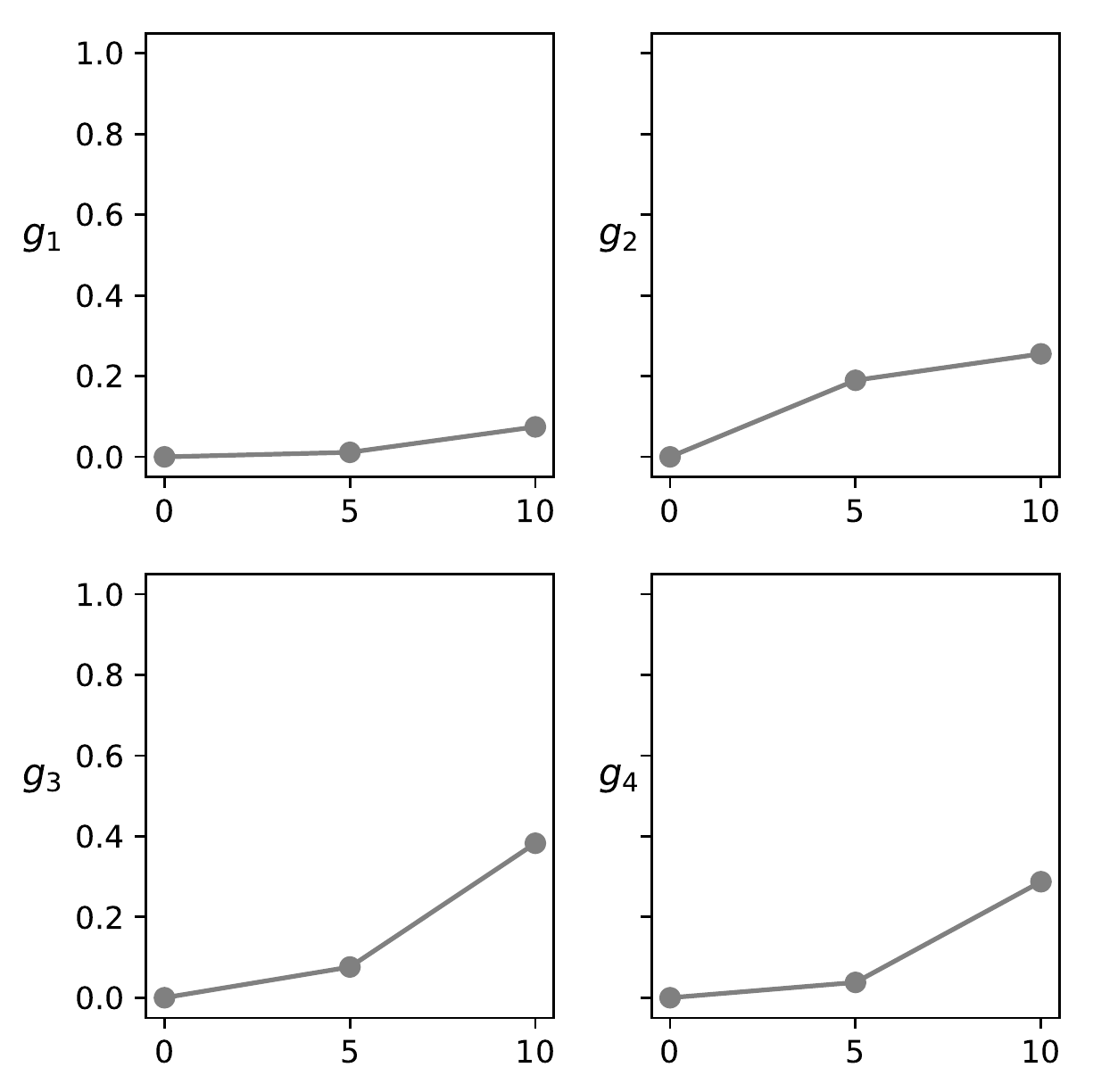}
\caption{Marginal value functions in the reference model.}\label{rys:case-study-reference-mvf}
\end{figure}

\begin{figure}[t]
\centering\includegraphics[width=12.5cm]{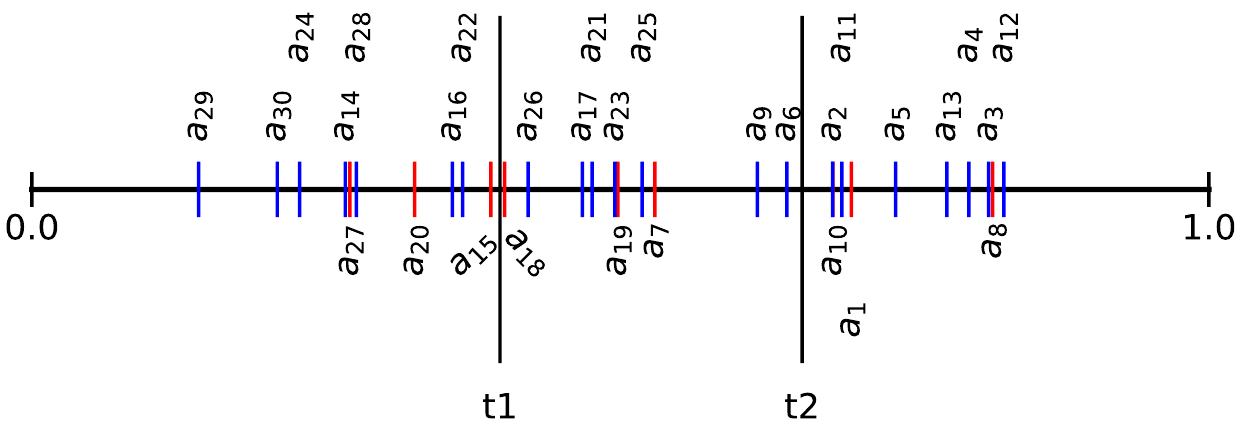}
\caption{Comprehensive values for all alternatives in relation to class thresholds in the reference model.}\label{rys:case-study-reference-global-values}
\end{figure}

 Then, we drew three reference alternatives to form the DM's indirect preference supplied as the input for UTADIS: $a_{15}, a_{20}, a_{27} \rightarrow C_{1}$, $a_{7}, a_{18}, a_{19} \rightarrow C_{2}$, and $a_{1}, a_{8}, a_{10} \rightarrow C_{3}$. To simulate the DM's policy, we randomly selected an additive value function with marginal functions depicted in Figure \ref{rys:case-study-reference-mvf}. All alternatives were assessed given this function (see Table \ref{tab:case_study_dataset} for marginal and comprehensive values). Subsequently, a pair thresholds ($t_{1} = 0.3977$ and $t_{2} = 0.6543$) was selected to delimit the three preference-ordered classes, and derive the assignments with the DM's reference model (see Table~\ref{tab:case_study_dataset}). Each class received ten alternatives. These alternatives are marked in red and their labels are provided under the axis in Figure~\ref{rys:case-study-reference-global-values}. Some procedures discussed in the previous section make use of robust results. In particular, we employed HAR for deriving $CAI'$s (see Table~\ref{tab:case_study_dataset}) and $APWI'$s (see~Table \ref{tab:case_study_apwi}). They were computed based on $10,000$ compatible sorting models. 
\begin{table}[h]
	\caption{Part of the matrix with the $APWI'$ values.}\label{tab:case_study_apwi}\centering\footnotesize
	\begin{tabular} {|l|l|l|l|l|l|l|l|l|l|}
		\hline
		\backslashbox{\boldsymbol{$a_{i}$}}{\boldsymbol{$a_{j}$}} & \textbf{\ldots} & \boldsymbol{$a_{9}$} & \boldsymbol{$a_{10}$} & \boldsymbol{$a_{11}$} & \textbf{\ldots} & \boldsymbol{$a_{24}$} & \boldsymbol{$a_{25}$} & \boldsymbol{$a_{26}$} & \textbf{\ldots} \\ \hline
		\textbf{\ldots} & \ldots & \ldots & \ldots & \ldots & \ldots & \ldots & \ldots & \ldots & \ldots \\ \hline
		\boldsymbol{$a_{9}$} & \ldots & 0.000 & 0.000 & 0.171 & \ldots & 1.000 & 0.725 & 0.880 & \ldots \\ \hline
		\boldsymbol{$a_{10}$} & \ldots & 0.124 & 0.000 & 0.200 & \ldots & 1.000 & 0.849 & 0.999 & \ldots \\ \hline
		\boldsymbol{$a_{11}$} & \ldots & 0.095 & 0.000 & 0.000 & \ldots & 1.000 & 0.649 & 0.800 & \ldots \\ \hline
		\textbf{\ldots} & \ldots & \ldots & \ldots & \ldots & \ldots & \ldots & \ldots & \ldots & \ldots \\ \hline
		\boldsymbol{$a_{24}$} & \ldots & 0.000 & 0.000 & 0.000 & \ldots & 0.000 & 0.000 & 0.000 & \ldots \\ \hline
		\boldsymbol{$a_{25}$} & \ldots & 0.000 & 0.000 & 0.000 & \ldots & 1.000 & 0.000 & 0.158 & \ldots \\ \hline
		\boldsymbol{$a_{26}$} & \ldots & 0.000 & 0.000 & 0.000 & \ldots & 0.992 & 0.000 & 0.000 & \ldots \\ \hline
		\textbf{\ldots} & \ldots & \ldots & \ldots & \ldots & \ldots & \ldots & \ldots & \ldots & \ldots \\ \hline
	\end{tabular}
\end{table}

\begin{figure}[t]
	\centering\includegraphics[width=15cm]{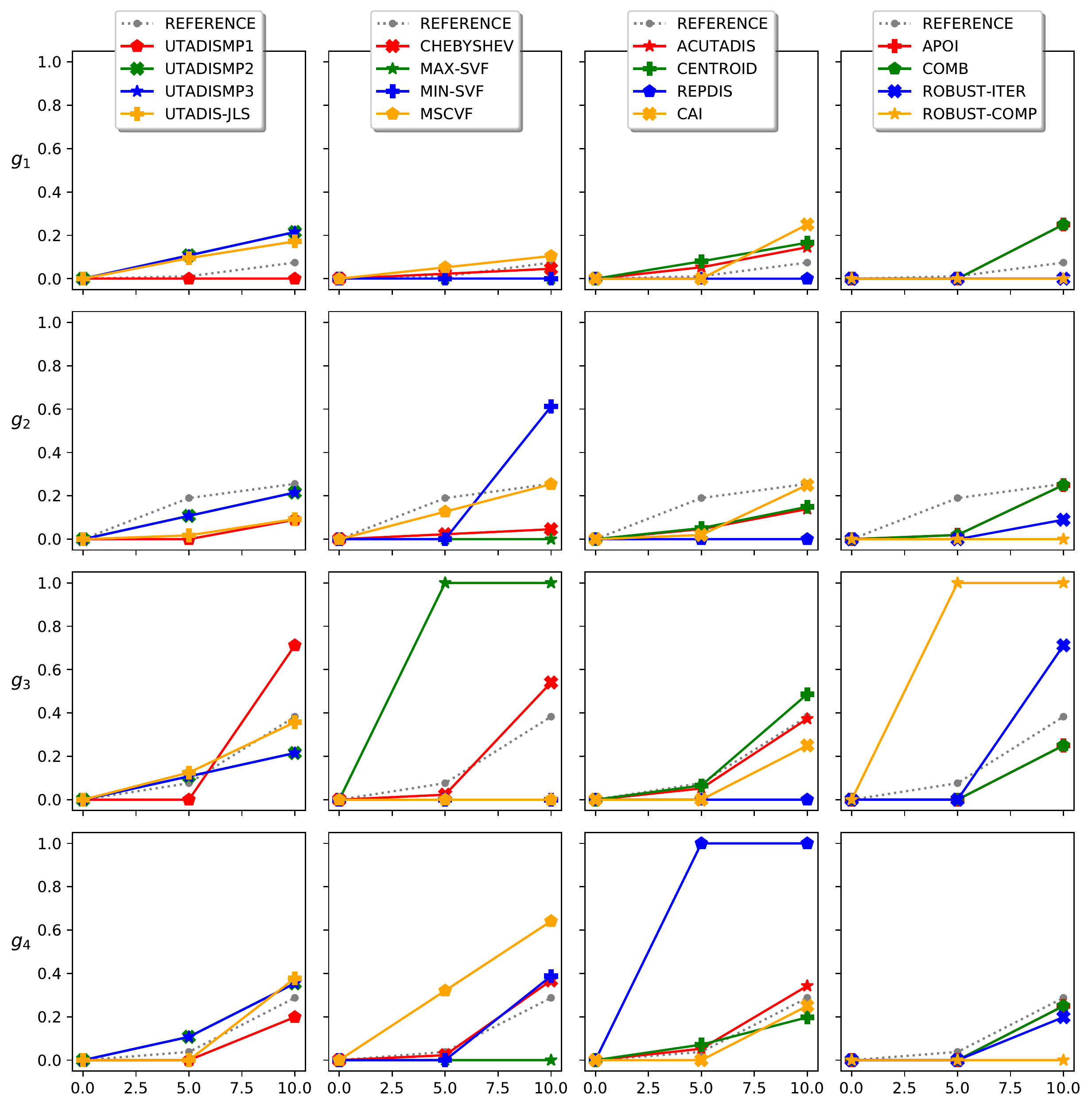}
	\caption{Marginal value function obtained with 16 procedures for selecting a representative sorting model.}\label{rys:case-study-methods-mvf}
\end{figure}

\begin{table}[h]
	\caption{Marginal values assigned to characteristic points and class thresholds obtained 16 procedures for selecting a representative sorting model. }\label{tab:case_study_ch_points}\centering \footnotesize
	\begin{tabular} {|l||l|l||l|l||l|l||l|l||l|l|}
		\hline
		\textbf{Method} & \boldsymbol{$u_{1}(5)$} & \boldsymbol{$u_{1}(10)$} & \boldsymbol{$u_{2}(5)$} & \boldsymbol{$u_{2}(10)$} & \boldsymbol{$u_{3}(5)$} & \boldsymbol{$u_{3}(10)$} & \boldsymbol{$u_{4}(5)$} & \boldsymbol{$u_{4}(10)$} & \boldsymbol{$t_{1}$}  & \boldsymbol{$t_{2}$}\\ \hline
		\textbf{REFERENCE} & \textbf{0.0113} & \textbf{0.0744} & \textbf{0.1898} & \textbf{0.2553} & \textbf{0.0762} & \textbf{0.3829} & \textbf{0.0380} & \textbf{0.2875}& \textbf{0.397722} & \textbf{0.654317} \\ \hline
		UTADISMP1 & 0.0000 & 0.0000 & 0.0000 & 0.0897 & 0.0000 & 0.7118 & 0.0000 & 0.1985 & 0.281516 & 0.408793 \\ \hline
		UTADISMP2 & 0.1076 & 0.2153 & 0.1076 & 0.2153 & 0.1076 & 0.2153 & 0.1076 & 0.3542 & 0.477009 & 0.678257 \\ \hline
		UTADISMP3 & 0.1076 & 0.2153 & 0.1076 & 0.2153 & 0.1076 & 0.2153 & 0.1076 & 0.3542 & 0.477009 & 0.678257 \\ \hline
		UTADIS-JLS & 0.0948 & 0.1722 & 0.0177 & 0.0927 & 0.1250 & 0.3571 & 0.0000 & 0.3779 & 0.399624 & 0.603032 \\ \hline
		CHEBYSHEV & 0.0229 & 0.0459 & 0.0229 & 0.0459 & 0.0229 & 0.5405 & 0.0229 & 0.3678 & 0.328974 & 0.500873 \\ \hline
		MAX-SVF & 0.0000 & 0.0000 & 0.0000 & 0.0000 & 1.0000 & 1.0000 & 0.0000 & 0.0000  & 0.999985 & 0.999988 \\ \hline
		MIN-SVF & 0.0000 & 0.0000 & 0.0000 & 0.6122 & 0.0000 & 0.0000 & 0.0000 & 0.3878 & 0.098490 & 0.245878 \\ \hline
		MSCVF & 0.0524 & 0.1047 & 0.1270 & 0.2540 & 0.0000 & 0.0000 & 0.3206 & 0.6413 & 0.514602 & 0.624040 \\ \hline
		ACUTADIS & 0.0527 & 0.1453 & 0.0466 & 0.1390 & 0.0519 & 0.3731 & 0.0527 & 0.3426& 0.352980 & 0.539950 \\ \hline
		CENTROID & 0.0801 & 0.1663 & 0.0511 & 0.1494 & 0.0654 & 0.4868 & 0.0715 & 0.1975 & 0.419456 & 0.591524 \\ \hline
		REPDIS & 0.0000 & 0.0000 & 0.0000 & 0.0000 & 0.0000 & 0.0000 & 1.0000 & 1.0000 & 0.999978 & 0.999985 \\ \hline
		CAI & 0.0000 & 0.2500 & 0.0195 & 0.2500 & 0.0000 & 0.2500 & 0.0000 & 0.2500 & 0.148776 & 0.354549 \\ \hline
		APOI & 0.0000 & 0.2500 & 0.0195 & 0.2500 & 0.0000 & 0.2500 & 0.0000 & 0.2500 & 0.148776 & 0.354549 \\ \hline
		COMB & 0.0000 & 0.2500 & 0.0195 & 0.2500 & 0.0000 & 0.2500 & 0.0000 & 0.2500 & 0.148776 & 0.354549 \\ \hline
		ROBUST-ITER & 0.0000 & 0.0000 & 0.0000 & 0.0897 & 0.0000 & 0.7118 & 0.0000 & 0.1985& 0.232163 & 0.359440 \\ \hline
		ROBUST-COMP & 0.0000 & 0.0000 & 0.0000 & 0.0000 & 1.0000 & 1.0000 & 0.0000 & 0.0000  & 0.999984 & 0.999986 \\ \hline
	\end{tabular}
\end{table}

\begin{table}[h]
	\caption{Comprehensive values for a subset of non-reference alternatives assigned by 16 procedures for selecting a representative sorting model.}\label{tab:case_study_global_values}\centering\footnotesize
%	\Rotatebox{90}{%
		\begin{tabular} {|l||l|l|l|l|l|l|l|l|l|}
			\hline
			\textbf{Method} & \boldsymbol{$a_{6}$} & \boldsymbol{$a_{9}$} & \boldsymbol{$a_{11}$} & \boldsymbol{$a_{17}$} & \boldsymbol{$a_{21}$} & \boldsymbol{$a_{22}$} & \boldsymbol{$a_{23}$} & \boldsymbol{$a_{25}$} & \boldsymbol{$a_{26}$} \\ \hline
			\textbf{REFERENCE} & \textbf{0.6414} & \textbf{0.6163} & \textbf{0.6881} & \textbf{0.4677} & \textbf{0.4759} & \textbf{0.3659} & \textbf{0.4952} & \textbf{0.5185} & \textbf{0.4216} \\ \hline
			UTADISMP1 & 0.5737 & 0.5542 & 0.5622 & 0.3348 & 0.4011 & 0.0871 & 0.4775 & 0.5342 & 0.4585 \\ \hline
			UTADISMP2 & 0.6449 & 0.6149 & 0.7139 & 0.4909 & 0.4898 & 0.4525 & 0.5231 & 0.4966 & 0.4442 \\ \hline
			UTADISMP3 & 0.6449 & 0.6149 & 0.7139 & 0.4909 & 0.4898 & 0.4525 & 0.5231 & 0.4966 & 0.4442 \\ \hline
			UTADIS-JLS & 0.5746 & 0.5151 & 0.6858 & 0.3643 & 0.3943 & 0.2763 & 0.5080 & 0.4574 & 0.4170 \\ \hline
			CHEBYSHEV & 0.5891 & 0.5358 & 0.6567 & 0.3475 & 0.4016 & 0.1578 & 0.5146 & 0.5172 & 0.4488 \\ \hline
			MAX-SVF & 1.0000 & 1.0000 & 1.0000 & 1.0000 & 1.0000 & 1.0000 & 1.0000 & 1.0000 & 1.0000 \\ \hline
			MIN-SVF & 0.1334 & 0.1296 & 0.2862 & 0.0256 & 0.0465 & 0.1207 & 0.1791 & 0.1008 & 0.0892 \\ \hline
			MSCVF & 0.6311 & 0.5940 & 0.7478 & 0.5181 & 0.5131 & 0.5314 & 0.5704 & 0.5228 & 0.4732 \\ \hline
			ACUTADIS & 0.5805 & 0.5399 & 0.6433 & 0.3661 & 0.3965 & 0.2548 & 0.4761 & 0.4644 & 0.4092 \\ \hline
			CENTROID & 0.6556 & 0.6456 & 0.6417 & 0.4639 & 0.4909 & 0.3120 & 0.5106 & 0.5438 & 0.4821 \\ \hline
			REPDIS & 1.0000 & 1.0000 & 1.0000 & 1.0000 & 1.0000 & 1.0000 & 1.0000 & 1.0000 & 1.0000 \\ \hline
			CAI & 0.4222 & 0.3910 & 0.4630 & 0.1488 & 0.1788 & 0.0930 & 0.2603 & 0.2472 & 0.2091 \\ \hline
			APOI & 0.4222 & 0.3910 & 0.4630 & 0.1488 & 0.1788 & 0.0930 & 0.2603 & 0.2472 & 0.2091 \\ \hline
			COMB & 0.4222 & 0.3910 & 0.4630 & 0.1488 & 0.1788 & 0.0930 & 0.2603 & 0.2472 & 0.2091 \\ \hline
			ROBUST-ITER & 0.5737 & 0.5542 & 0.5622 & 0.3348 & 0.4011 & 0.0871 & 0.4775 & 0.5342 & 0.4585 \\ \hline
			ROBUST-COMP & 1.0000 & 1.0000 & 1.0000 & 1.0000 & 1.0000 & 1.0000 & 1.0000 & 1.0000 & 1.0000 \\ \hline
		\end{tabular}
%	}%
	
\end{table}

\begin{table}[h]
	\caption{Class assignments for a subset of non-references alternatives determined with 16 procedures for selecting a representative sorting model.}\label{tab:case_study_assignments}\centering\footnotesize
	\begin{tabular} {|l||l|l|l|l|l|l|l|l|l|}
		\hline
		\textbf{Method} & \boldsymbol{$a_{6}$} & \boldsymbol{$a_{9}$} & \boldsymbol{$a_{11}$} & \boldsymbol{$a_{17}$} & \boldsymbol{$a_{21}$} & \boldsymbol{$a_{22}$} & \boldsymbol{$a_{23}$} & \boldsymbol{$a_{25}$} & \boldsymbol{$a_{26}$} \\ \hline
		\textbf{REFERENCE} & \textbf{2} & \textbf{2} & \textbf{3} & \textbf{2} & \textbf{2} & \textbf{1} & \textbf{2} & \textbf{2} & \textbf{2} \\ \hline
		UTADISMP1 & 3 & 3 & 3 & 2 & 2 & 1 & 3 & 3 & 3 \\ \hline
		UTADISMP2 & 2 & 2 & 3 & 2 & 2 & 1 & 2 & 2 & 1 \\ \hline
		UTADISMP3 & 2 & 2 & 3 & 2 & 2 & 1 & 2 & 2 & 1 \\ \hline
		UTADIS-JLS & 2 & 2 & 3 & 1 & 1 & 1 & 2 & 2 & 2 \\ \hline
		CHEBYSHEV & 3 & 3 & 3 & 2 & 2 & 1 & 3 & 3 & 2 \\ \hline
		MAX-SVF & 3 & 3 & 3 & 2 & 2 & 1 & 3 & 3 & 3 \\ \hline
		MIN-SVF & 2 & 2 & 3 & 1 & 1 & 2 & 2 & 2 & 1 \\ \hline
		MSCVF & 3 & 2 & 3 & 2 & 1 & 2 & 2 & 2 & 1 \\ \hline
		ACUTADIS & 3 & 2 & 3 & 2 & 2 & 1 & 2 & 2 & 2 \\ \hline
		CENTROID & 3 & 3 & 3 & 2 & 2 & 1 & 2 & 2 & 2 \\ \hline
		REPDIS & 2 & 2 & 2 & 2 & 2 & 1 & 2 & 2 & 2 \\ \hline
		CAI & 3 & 3 & 3 & 2 & 2 & 1 & 2 & 2 & 2 \\ \hline
		APOI & 3 & 3 & 3 & 2 & 2 & 1 & 2 & 2 & 2 \\ \hline
		COMB & 3 & 3 & 3 & 2 & 2 & 1 & 2 & 2 & 2 \\ \hline
		ROBUST-ITER & 3 & 3 & 3 & 2 & 3 & 1 & 3 & 3 & 3 \\ \hline
		ROBUST-COMP & 3 & 3 & 3 & 1 & 2 & 1 & 3 & 3 & 2 \\ \hline
	\end{tabular}
\end{table}

In what follows, we discuss the results obtained with 16 procedures for selecting a representative sorting model. The respective MVFs are illustrated in Figure~\ref{rys:case-study-methods-mvf}. For precise marginal values assigned to the characteristic points and class thresholds, see Table~\ref{tab:case_study_ch_points}. Tables~\ref{tab:case_study_global_values} and~\ref{tab:case_study_assignments} show the comprehensive values and class assignments determined with all approaches. To save space, we provide detailed results only for nine non-reference cities for which at least one method recommended a class that differed from the one assigned by the reference model. For the remaining twelve alternatives, all 16 procedures recommended an assignment compatible with the indication of the reference model. 

The UTADISMP1 method aims at reproducing the DM's preferences by maximizing the difference between comprehensive values of reference alternatives and the thresholds of their desired classes. This objective implies that due to the existence of reference alternatives with comprehensive values close to the thresholds (e.g., $a_{15}$ and $a_{18}$), the resulting marginal function and class thresholds differ from the reference ones. The operational procedure underlying UTADISMP1 implies that non-reference alternatives with very similar performance profiles to reference alternatives are assigned to the same class. This can be observed for, e.g., $a_{19}$ and $a_{17}$ or $a_{6}$ and $a_{10}$.

The models obtained with UTADISMP2 and UTADISMP3 are the same. This is understandable since both procedures account for maximizing the minimal slope of MVFs, while UTADISMP2 additionally considers the same objective of UTADISMP1. The evidence of maximizing the differences between marginal values assigned to successive characteristic points is visible in Table \ref{tab:case_study_ch_points}. For all criteria, $u_{j}(5)$ has the same value ($0.10763$), and in three cases, $u_{j}(10)$ is exactly twice as large (0.21526), hence satisfying the monotonicity constraints with a large margin ($\rho$). In this case, the slacks for other constraints were rather marginal. For example, comprehensive values of two reference alternatives $U(a_{15}) = 0.4770076$ and $U(a_{18}) = 0.4770096$ are very close to threshold $t_1 = 0.4770086$, though being assigned to different classes: $C_1$ and $C_2$, respectively. A characteristic consequence of maximizing $\rho$ is that for many problems, the solutions obtained by these two methods have a relatively even distribution of the maximal values of MVFs and their curvatures are close to being linear. 

An explicit mechanism for deriving the marginal functions which minimally deviate from linearity is implemented in MSCVF. For the considered problem, it obtained an ideal model, satisfying the following condition: $\forall_{j \in \{1, 2, 3, 4\}} \frac{u_{j}(10) - u_{j}(5)}{10 - 5} = \frac{u_{j}(5) - u_{j}(0)}{5 - 0}$ for all criteria, which translated to the lowest possible objective function's value ($\phi = 0$). The linear MVFs are visible in Figure~\ref{rys:case-study-methods-mvf}. Obviously, attaining such a parsimony is not possible for all problems as it depends on the alternatives' performances and reference assignments. 

The MAX-SVF procedure maximizes the sum of comprehensive values of reference alternatives. In the presence of scarce preference information, when the performances on a single criterion align with the order of desired classifications, this objective may lead to one MVF assigning values close to one for all characteristic points (except the one corresponding to the least preferred performance). For the considered study, this is observed for $u_3$ with $u_{3}(5)= 0.9999797$ and $u_{3}(10) = 0.9999942$, whereas the impact of the remaining criteria is negligible ($u_{1}(10) = 0.0$, $u_{2}(10) = 1.82 \times 10^{-6}$, and $u_{4}(10) = 4.02 \times 10^{-6}$). Most probably, such a model has few chances for adequately representing the comprehensive DM's policy of quality assessment. Nonetheless, it may reproduce the reference class assignments relatively well (16 out of 21 non-references alternatives were classified correctly). All misclassified non-reference alternatives are assigned to more preferred classes than the reference ones (e.g., $a_{26}$ is assigned to $C_3$, while the reference model suggested $C_2$).

The MIN-SVF procedure minimizes the sum of comprehensive values of reference alternatives. While MAX-SVF is constrained to assign $U$ equal to zero to an anti-ideal alternative, MIN-SVF needs to oppose its objective against the requirement of assigning $U$ equal to one to an ideal alternative. For the considered problem, this is attained by assigning large positive values to the best performances on $g_2$ ($u_{2}(10) = 0.6122$) and $g_4$ ($u_{4}(10) = 0.3878$). However, the marginal values of all remaining characteristic points are equal to zero, leading to low comprehensive values of alternatives. As a result, the threshold values are also the lowest among all methods (see $t_{1} = 0.098490$ and $t_{2} = 0.245878$ in Table \ref{tab:case_study_ch_points}). Unlike for MAX-SVF, the class recommended for all misclassified non-reference alternatives is always less preferred than the reference one (e.g., $a_{26}$ is assigned to $C_1$, while the reference assignment is $C_2$).

UTADIS-JLS is a heuristic approach that constructs a representative model by averaging the extreme compatible ones that maximize and minimize the greatest value of the individual MVFs. For the considered problem, this led to non-negligible maximal shares of all MVFs with the predominant role of $g_3$ ($u_{3}(10) = 0.3571$) and $g_4$ ($u_{4}(10) = 0.3779$) and well-distributed class thresholds ($t_1 = 0.3999624$ and $t_2 = 0.603032$). Interestingly, in the final model, $u_{4}(5) = 0.0$, which means that in eight intermediate model this marginal value was also equal to zero. Analyzing the results obtained with other methods, many solutions repeat this pattern. This suggests that low scores (below 5.0) w.r.t. waste and land use ($g_4$) may have no or negligible impact on the recommended class assignments. As a result, Kiev ($g_{4}(a_{30}) = 1.43$) and Istanbul ($g_{4}(a_{16}) = 4.86$) are often scored equally on $u_4$, despite a noticeable difference in their performances. 

The CENTROID method is similar to UTADIS-JLS in terms of deriving an average model. However, when doing so, it considers a large sample of uniformly distributed models. The marginal value functions obtained with CENTROID confirm that the extreme models considered by UTADIS-JLS are not representative for the entire feasible polyhedron. In particular, the maximal shares of $u_2$ and $u_3$ are greater, whereas the impact of $u_4$ is reduced. A detailed analysis of the derived model confirms that incomplete indirect preference information (in this case, concerning $9$ out of $30$ alternatives) does not allow reproducing the reference model accurately, even if the assignment examples are perfectly reproduced. When comparing the two models in Figure~\ref{rys:case-study-methods-mvf}, one can observe the overestimation of the maximum value for $g_{1}$ ($u_{1_{REFERENCE}}(10) = 0.0744$ and $u_{1_{CENTROID}}(10) = 0.1663$) and $g_{3}$ ($u_{3_{REFERENCE}}(10) = 0.3829$ and $u_{3_{CENTROID}}(10) = 0.4868$) and the underestimation for $g_{2}$ ($u_{2_{REFERENCE}}(10) = 0.2553$ and $u_{2_{CENTROID}}(10) = 0.1494$) and $g_{4}$ ($u_{4_{REFERENCE}}(10) = 0.2875$ and $u_{4_{CENTROID}}(10) = 0.1975$).

In the CHEBYSEV method, the ``central" model is determined in a more formalized way as the center of the hypersphere inscribed in the polyhedron defining the set of all compatible sorting models. For this purpose, the constraints incorporate variable $r$ representing the value of hypersphere radius. The obtained MVFs are similar to those obtained with UTADISMP2 in the sense of assigning the same marginal values to mid-points on all criteria ($u_{j}(5)$). Also, the values assigned to the end points $u_{1}(10)$ and $u_{2}(10)$ are exactly twice as large. This is due to optimizing variable $r$, which is responsible for maximizing the minimal differences between marginal values assigned to successive characteristic points. In addition, this variable is also used in constraints reproducing the class assignments as the hypersphere radius depends on these constraints too. As a result, the comprehensive values of reference alternatives also highly diverge from the thresholds, which is mainly attained thanks to high maximal shares on $u_3$ and $u_4$. 

In the same spirit, ACUTADIS derives a central model corresponding to an analytic center of the polyhedron. The underlying optimization model is non-linear, considering the sum of logarithms of the slack variables involved in each inequality. The obtained MVFs are strictly increasing, the class thresholds are well-separated, and $u_3$ and $u_4$ have about $2.5$ times greater impact on the comprehensive values than $u_1$ and $u_2$. ACUTADIS is also one out of only four methods that made only a single mistake in classifying the non-reference alternatives. The incorrectly rated city is Paris ($a_{6}$), which is relatively similar to London ($a_{10}$) assigned by the DM to $C_3$. The latter alternative is distant from the lower threshold of its desired class ($U(a_{10}) = 0.6143$ and  $t_{2} = 0.5399$). This implies that the comprehensive value of Paris also fits in the range associated with the most preferred class.

The REPDIS procedure returned a model which builds the comprehensive scores based on just a single criterion (in this case -- $g_{4}$). Hence, the maximal share of $u_4$ is equal to one, whereas the marginal value assigned to $u_4(5)$ is very close to one (0.999974). Let us emphasize that the numbers provided in Tables~\ref{tab:case_study_ch_points} and~\ref{tab:case_study_global_values} are rounded to four decimal places. As a result, the differences between comprehensive values of a large set of alternatives as well as class thresholds are extremely small. This is an undesired effect from the viewpoint of results' interpretability. It suggests that for this particular problem, the objectives built on the analysis of $APWI'$s proved too challenging to let the method emphasize the value differences for all pairs of alternatives simultaneously. To maximize its objective function, while taking into account the conflicting sub-objectives, REPDIS opted for balancing the alternatives' comprehensive assessments. The same problem can be observed for ROBUST-COMP with the proviso that in this case, criterion $g_3$ was used as the sole one from which alternatives derived positive values. A side effect of such small differences is that when comparing the classification suggested by such models for non-reference alternatives and the ones derived with the DM's simulated model, there is no match for many pairs. In the case of ROBUST-COMP such mistakes are observed for $5$ out of $21$ cities. 

ROBUST-ITER and ROBUST-COMP take into account the necessary assignment-based preference relations. While ROBUST-COMP attempts to consider the two objectives relevant for this approach at once, ROBUST-ITER optimizes them one after another. For the illustrative study, such an approach led to a more intuitive and interpretable model. In fact, the model obtained after considering the first objective was not modified when subsequently optimizing the other objective. Hence, the resulting model was determined solely by maximizing the value differences for pairs of alternatives related by the necessary assignment-based preference relation (e.g., ($a_7, a_{15}$) among reference alternatives and ($a_{25}, a_{24}$) among non-reference alternatives). The value differences for pairs always assigned to the same class were just a side effect of the primary optimization. Clearly, this observation does not hold for all decision problems because the secondary objective can often break ties when selecting among models that optimize the primary objective equally well. When it comes to the assignments of non-reference alternatives, ROBUST-ITER misclassified 6 out of 21 cities compared to the assignments provided by the DM's reference model.

The three novel approaches proposed in this paper (CAI, APOI, and COMB) selected the same model for the considered problem. COMB, putting the objective functions of CAI and APOI together, often returns a result that matches the solution of either model. However, such a~perfect agreement between CAI and APOI is less common. Nevertheless, it can be easily justified because they build their outcomes on the stochastic acceptability indices, even if CAI focuses on the class assignments and APOI considers assignment-based pairwise relations. The model discovered by these approaches is characterized by equal maximal shares of all criteria ($0.25$) and a~positive marginal value assigned to the mid-point only for $u_2$. Such a balanced distribution implied relatively low comprehensive values of all alternatives (see Table~\ref{tab:case_study_global_values}) and low thresholds separating the classes ($t_1 = 0.148776$ and $t_2 = 0.354549$). To explain the operational procedure of CAI and APOI, let us focus on Riga and Athens. According to \ac{SOR}, Riga is assigned to $C_1$ by all models ($CAI'(a_{15}, C_{1}) = 1.0$). For Athens, there is an ambiguity in the assignments ($CAI'(a_{17}, C_{1}) = 0.088$ and $CAI'(a_{17}, C_{2}) = 0.912$). As a result, they are assigned to a class better than Riga for the vast majority ($91.2\%$) of models ($APOI'(a_{15}, a_{17}) = 0.088$ and $APOI'(a_{17}, a_{15}) = 1$). Hence to optimize the objective functions' values and emphasize the most frequent results in the representative models, the novel procedures opt for assigning Riga to $C_1$ and Athens to $C_2$, even if it was challenging to separate these two alternatives ($U(a_{15}) = 0.1487745$, $U(a_{17}) = 0.1487765$, and $t_{1} = 0.1487755$).

\section{Computational experiments}
\label{sec:experiment}
\noindent This section is devoted to the computational experiments performed to examine the quality of procedures for selecting a representative sorting model. We define the measures used to compare the 16 approaches and the characteristics of problems instances considered during the tests. The results obtained for each measure are discussed in detail, given the average outcomes across all considered settings and performance trends observed when changing some parameter values.

\subsection{Comparative measures}
\label{sec:measures}
\noindent The performance of procedures for selecting a single sorting model will be quantified in terms of five measures. On the one hand, they concern how the assignments and models used to simulated the decision-making policy are reflected in the delivered results based on incomplete preference information. On the other hand, they reflect how representative are the recommended assignment for the entire set of compatible sorting models. 

Let us denote a set of all non-reference alternatives that the DM has not classified by $A^{T} = A \setminus A^{R}$. The reference model composed of marginal value functions, comprehensive values, and class thresholds is denoted by $U^{REF}$, and the analogous model returned by procedure $P$ is $U^{P}$. Finally, the assignment conducted by model $U$ is denoted by $\rightarrow^{U}$.

\noindent \textbf{Classification accuracy.} 
To determine the quality of the proposed sorting model, we can verify how far the model proposed by the procedure is from the comprehensive DM's preferences in terms of recommended assignments. We focus only on the non-reference alternatives because all procedures reproduce the assignments of reference solutions. Therefore, the classification accuracy captures the proportion of alternatives in set $A^T$ for which the recommended and reference assignments agree, i.e.~\cite{Doumpos2014}, : 
\begin{equation}
accuracy(U^{P}) = \frac{|a: a \in A^{T} \land a \rightarrow^{U^{P}} C_{l_{P}} \land a \rightarrow^{U^{REF}} C_{l_{REF}} \land l_{P} = l_{REF}|}{|A^{T}|}.
\end{equation}
For example, when considering the results reported in Table~\ref{tab:case_study_assignments} for 9 non-reference alternatives and remembering that the classification of the remaining 12 test options agreed with the references one, the classification accuracy obtained by UTADISMP2 was $accuracy(U^{UTADISMP2}) = 20/21 = 0.9524$ and for CHEBYSHEV -- it was $accuracy(U^{CHEBYSHEV}) = 17/21 = 0.8095$. The former procedure misclassified only $a_{26}$, whereas for the latter -- four non-reference alternatives ($a_{6}$, $a_{9}$, $a_{23}$, and $a_{25}$) were classified incorrectly w.r.t. the reference assignment. 

\label{subsec:assignments_acceptability}
\noindent \textbf{Assignment acceptability.} Another measure compares the assignments recommended by different procedures with the classification obtained in the entire set of sorting models. The assignment acceptability reflects average support given to the assignments recommended by a given procedure for all non-reference alternatives in terms of class acceptability indices $CAI'$s derived from the analysis of all feasible solutions, i.e.:
\begin{equation}
MCAI(U) = \frac{1}{|A^{T}|} \sum_{a \in A^{T}: a \rightarrow^{U} C_{l_{a}}} CAI'(a, C_{l_{a}}).
\end{equation}
The maximal $MCAI$ value can be obtained when each non-reference alternative is assigned to the class with the highest $CAI$ value. As noted in \cite{Doumpos2014}, this approach to classification is based on the \emph{robust assignment rule}. The value defined in this way is marked as $MCAI_{max}$:
\begin{equation}
MCAI_{max} = \frac{1}{|A^{T}|} \sum_{a \in A^{T}} \max_{l \in \{1, \ldots, p\}} CAI'(a, C_{l}).
\end{equation}
In what follows, we consider an \emph{Absolute MCAI} ($MCAI_{abs} = MCAI(U^{P})$), and the \emph{Relative MCAI}, which makes the measure values more interpretable by referring them to the best possible solution that could be obtained for a~given problem:
\begin{equation}
MCAI_{rel}(U^{P}) = \frac{MCAI(U^{P}) - MCAI_{max}}{MCAI_{max}} = \frac{MCAI(U^{P})}{MCAI_{max}} - 1.
\end{equation}
When considering $CAI'$s reported in Table~\ref{tab:case_study_dataset}, for 21 non-reference alternatives, $MCAI_{max}$ is equal to $0.9656$. In fact, the maximal $CAI'$ was lesser than one only for 9 alternatives. Four approaches (CENTROID, CAI, APOI, and COMB) identified a solution with $MCAI_{abs} = MCAI_{max}$. Consequently, for the methods, $MCAI_{rel}(U^{P}) = 0$. Hence these procedures perfectly reflect the most robust assignments. Note that this value is lower for the reference model, which assigns alternatives $a_{6}$ and $a_{9}$ to class $C_{2}$. However, their class acceptabilities for class $C_3$ are higher than for $C_2$ (e.g., for $a_6$ -- $CAI'(a_{6}, C_{2}) = 0.056$ and $CAI'(a_{6}, C_{3}) = 0.944$). As a result, for the reference model, $MCAI_{abs}(U^{REF}) = 0.8875$ and $MCAI_{rel}(U^{REF}) = \frac{0.8875}{0.9656} - 1 = -0.0809$. This example emphasizes that $MCAI$ captures whether a given procedure reconstructs the most common results observed for all compatible sorting model rather than reconstruction of the reference assignments. 

\noindent The following three measures focus on the similarity between models rather than assignments.

\noindent \textbf{Differences between marginal values.} To capture the agreement between shapes of MVFs, we compare the marginal values assigned to all characteristic points except the least preferred. The latter ones are, by definition, always assigned values equal to zero. Such a measure -- summarizing absolute value differences -- can be considered as the comprehensive distance between the reference model and $U^{P}$ obtained with procedure $P$:
\begin{equation}
\Delta^{REF}_{TO}(U^{P}) = \frac{1}{\sum\limits_{j = 1}^{m} (n^{A}_{j} - 1)} \sum_{j = 1}^{m} \sum_{k = 2}^{n^{A}_{j}} |u^{U^{P}}_{j}(\beta^{k}_{j}) - u^{U^{REF}}_{j}(\beta^{k}_{j})|.
\end{equation}
Another perspective concerns the distance of MVFs from a sorting model that represents well the feasible space of all models. For this purpose, we adopt the outcomes of the CENTROID procedure, which is an average of a large sample of uniformly distributed value functions and class thresholds. It can be defined in the following way:
\begin{equation}
\Delta^{CENT}_{TO}(U^{P}) = \frac{1}{\sum\limits_{j = 1}^{m} (n^{A}_{j} - 1)} \sum_{j = 1}^{m} \sum_{k = 2}^{n^{A}_{j}} |u^{U^{P}}_{j}(\beta^{k}_{j}) - u^{U^{CENT}}_{j}(\beta^{k}_{j})|.
\end{equation}
When considering the results reported in Table \ref{tab:case_study_ch_points}, the model which is the closest to the reference one in terms of $\Delta^{REF}_{TO}$ was obtained with ACUTADIS ($\Delta^{REF}_{TO}(U^{ACUTADIS}) = 0.0595$). On the other extreme, REPDIS identified the furthest solution from the reference model ($0.3331$). As far as the comparison with an average model is concerned, the outcome of the CENTROID procedure is, by definition, the same (i.e., $\Delta^{CENT}_{TO}(U^{CENTROID}) = 0.0$). However, other methods which also aimed for identifying a central model attained quite favorable scores too: for ACUTADIS -- $\Delta^{CENT}_{TO}(U^{ACUTADIS}) = 0.0443$, for UTADIS-JLS -- $0.0690$, and for CHEBYSHEV -- $0.0781$. Again, for REPDIS, the distance was vast ($0.3413$). 

\noindent \textbf{Differences between comprehensive values.} Another measure refers to the aggregated results at the level that considers all criteria jointly. Instead of comparing the MVFs, it summarizes the differences between comprehensive values attained by all non-reference alternatives for the reference and resulting models: 
\begin{equation}
\Delta^{REF}_{CV}(U^{P}) = \frac{1}{|A^{T}|} \sum_{a \in A^{T}} |U^{U^{P}}(a) - U^{U^{REF}}(a)|.
\end{equation}
Part of the results needed to compute such values for the illustrative study is available in Table \ref{tab:case_study_global_values}. Taking into account the comprehensive values of 21 non-reference alternatives, the closest model to the reference one was obtained with CENTROID ($\Delta^{REF}_{CV}(U^{CENTROID}) = 0.0265$). In turn, the furthest distance can be attributed to MAX-SVF and ROBUST-COMP (in both cases, $\Delta^{REF}_{CV}(U^{P}) = 0.4287$).

\noindent \textbf{Differences between thresholds values.} The last measure concerns the similarity between separating class thresholds in the reference and resulting models:
\begin{equation}
\Delta^{REF}_{TH}(U^{P}) = \frac{1}{p-1} \sum_{l = 1}^{p-1} |t^{U^{P}}_l - t^{U^{REF}}_l|.
\end{equation}
It captures if the method can reproduce the range width of comprehensive values that justify an assignment to a~given class and their positions on the scale of AVF. 

For the illustrative example, its values can be determined based on Table~\ref{tab:case_study_ch_points}. For UTADIS-JLS, the threshold values are the closest to the reference model ($\Delta^{REF}_{TH}(U^{UTADIS-JLS}) = 0.0266$). On the other extreme, they are the furthest for MAX-SVF, REPDIS, and ROBUST-COMP (for all these procedures, $\Delta^{REF}_{TH}(U^{P}) = 0.4740$). Indeed, the separation between classes was very poor for these methods, and all thresholds were close to one.

\subsection{Experimental setting}
\noindent When generating instances of test problems, we followed~\cite{Doumpos2014} in considering various settings for the dimensionality of data:
\begin{itemize}
	\item the number of classes -- $p \in \{2, 3, 4, 5\}$;
	\item the number of criteria -- $m \in \{3, 5, 7, 9\}$;
	\item the number of characteristic points for each criterion $g_j$ -- $\gamma_j \in \{2, 4, 6\}$;
	\item the number of reference alternatives assigned by the DM to each of $p$ classes -- $R \in \{3, 5, 7, 10\}$.
\end{itemize}
In this way, we covered relatively simple problems with two linear criteria and six reference alternatives in two classes, and complex problems with $9$ criteria associated with marginal functions with $6-1=5$ linear pieces and up to $50$ reference alternatives in five decision classes. The number of non-reference alternatives from set $A^{T}$ is ten for each class. In this way, we represent the realistic scenarios in which the set of reference alternatives is at least as large as the test (non-reference) set. Consequently, the greatest problem instances involved up to $100$ alternatives. It is a high value when considering the typical MCDA setting, which nevertheless still makes feasible the execution of robustness analysis methods incorporated by some of the considered procedures. For each combination of parameter values, we averaged the results over $100$ problem instances. Hence we considered $4 \times 4 \times 3 \times 4 \times 100 = 19,200$ instances in total.

For each instance, we followed the procedure described in~\cite{Doumpos2014}. Hence two pools, each composed of $1,000$ alternatives, were generated. The reference alternatives were randomly selected from the first pool, and the test (non-reference) alternatives were chosen from the other pool. The alternatives in these two pools were evaluated with a randomly generated AVF serving as the DM's reference model. For simplicity, we assumed that the number of characteristic points for the respective MVFs was equal to $\gamma_j$ in the considered problem setting. Then, the separating class thresholds $t = [t_1, \ldots, t_{p-1}]$ were set to respect the following proportions of alternatives from the first (reference) pool being assigned to particular classes: for $p=2$ -- 50-50, for $p=3$ -- 30-40-30, for $p=4$ -- 20-30-30-20, and for $p=5$ -- 15-20-30-20-15. Such divisions correspond to realistic scenarios in which extreme classes are less common than intermediate ones. Thus specified thresholds were used to derive class assignments for alternatives contained in both pools. Finally, a pre-defined number of alternatives, depending on the considered setting, were randomly selected for each class to construct sets of reference and test alternatives. When put together, these two sets ($A^{R}$ and $A^{T}$) formed a set of alternatives $A$ that would be normally considered by the DM facing a particular decision problem.

The 16 methods were run for all problem instances except for MSCVF for problems with $\gamma_j=2$ characteristic points. In this case, the marginal value functions for all methods are linear. For each problem instance, the values of stochastic acceptability indices were estimated based on $10,000$ sorting models generated with \ac{HAR}~\cite{CiomekHAR}.

\subsection{Results}
\label{sec:results}
\noindent In this section, we discuss the results of an experimental comparison of the 16 procedures for selection of a single, representative sorting model. For each measure, we consider the outcomes averaged over all problem instances and the mean values of the performance measures obtained for different values of each problem dimension ($p$, $m$, $\gamma_j$, and $R$). 

\subsubsection{Classification accuracy}
\label{subsec:classification_accuracy_results}
\noindent Average classification accuracies over all problem instances are provided in Table \ref{tab:accuracy_all}. The difference between the best and worst performers is substantial (around $16\%$). The best accuracy was obtained by ACUTADIS ($0.8313$), which identifies an analytic center of the polyhedron using non-linear optimization. In general, seeking the central solution proves to be an excellent strategy to increase classification accuracy. This is confirmed by the results attained by other approaches implementing this concept, i.e., CENTROID ($0.8134$) and CHEBYSHEV ($0.8099$). Highly favorable results (between $0.8113$ and $0.8119$) are obtained by the approaches exploiting the stochastic acceptability indices: CAI, APOI, and COMB. The advantageous performance of these methods, along with the high position of CENTROID, confirms the usefulness of conducting robustness analysis with the Monte Carlo simulations. Slightly lesser classification accuracies were attained with the traditional procedures, which are most often used in the context of UTADIS due to their simplicity, i.e., UTADISMP1, UTADISMP2, and UTADIS-JLS. They choose either the most discriminant model or an average model, though, based on the analysis of extreme ones only. 

The worst accuracies were obtained by procedures optimizing the sum of comprehensive values (for MAX-SVF -- 0.6778 and for MIN-SVF -- 0.6752). Similarly, focussing only on the shape of MVFs, as done by UTADISMP3 and MSCVF, did not lead to high accuracies. Also, exploiting the exact outcomes of robustness analysis by ROBUST-ITER and ROBUST-COMP allowed for reproducing the correct assignment for over $10\%$ less non-reference alternatives than ACUTADIS. The objectives considered by these approaches differ vastly from the best-performing methods. A general conclusion from the experiment is that when one aims to maximize the classification accuracy, a sorting model should be selected by exploiting the feasible polyhedron or considering the robustness of shapes or recommendations delivered with a large subset of all compatible models.

\begin{table}[h]
	\caption{Mean values and standard deviations of classification accuracy for all considered problem settings, different numbers of classes and criteria.}\label{tab:accuracy_all}\centering\footnotesize
	\begin{tabular} {|l||l|l||l|l|l|l||l|l|l|l|}
		\hline
		& \multicolumn{2}{c||}{All settings} & \multicolumn{4}{c||}{Number of classes}  & \multicolumn{4}{c|}{Number of criteria}\\
		\hline
		\textbf{Procedure} & \textbf{mean} & \textbf{std}  & \textbf{2} & \textbf{3} & \textbf{4} & \textbf{5} & \textbf{3} & \textbf{5} & \textbf{7} & \textbf{9} \\ \hline
		UTADISMP1 & 0.7897 & 0.1252  & 0.8094 & 0.7852 & 0.7814 & 0.7829  & 0.8503 & 0.8023 & 0.7686 & 0.7376 \\ \hline
		UTADISMP2 & 0.7756 & 0.1210 & 0.8108 & 0.7760 & 0.7604 & 0.7553 & 0.8294 & 0.7872 & 0.7567 & 0.7292 \\ \hline
		UTADISMP3 & 0.7483 & 0.1231 & 0.8101 & 0.7494 & 0.7207 & 0.7132 & 0.7989 & 0.7569 & 0.7287 & 0.7088 \\ \hline
		UTADIS-JLS & 0.7703 & 0.1345 & 0.7917 & 0.7621 & 0.7606 & 0.7669 & 0.8237 & 0.7859 & 0.7506 & 0.7211 \\ \hline
		CHEBYSHEV & 0.8099 & 0.1124 & 0.8423 & 0.8074 & 0.7956 & 0.7942 & 0.8623 & 0.8204 & 0.7918 & 0.7650 \\ \hline
		MAX-SVF & 0.6778 & 0.1530 & 0.7441 & 0.6774 & 0.6479 & 0.6416 & 0.7747 & 0.6947 & 0.6419 & 0.5997 \\ \hline
		MIN-SVF & 0.6752 & 0.1515& 0.7424 & 0.6743 & 0.6467 & 0.6373 & 0.7701 & 0.6906 & 0.6378 & 0.6021 \\ \hline
		MSCVF & 0.7100 & 0.1304 & 0.7521 & 0.7097 & 0.6956 & 0.6828 & 0.7860 & 0.7234 & 0.6811 & 0.6496 \\ \hline
		ACUTADIS & 0.8313 & 0.1040  & 0.8548 & 0.8288 & 0.8214 & 0.8203  & 0.8794 & 0.8424 & 0.8136 & 0.7899 \\ \hline
		CENTROID & 0.8134 & 0.1140 & 0.8434 & 0.8129 & 0.7993 & 0.7979 & 0.8714 & 0.8255 & 0.7915 & 0.7650 \\ \hline
		REPDIS & 0.7571 & 0.1207 & 0.7841 & 0.7545 & 0.7444 & 0.7456 & 0.8138 & 0.7684 & 0.7349 & 0.7114 \\ \hline
		CAI & 0.8119 & 0.1145 & 0.8426 & 0.8118 & 0.7975 & 0.7958 & 0.8705 & 0.8242 & 0.7897 & 0.7632 \\ \hline
		APOI & 0.8113 & 0.1151 & 0.8412 & 0.8114 & 0.7971 & 0.7952 & 0.8695 & 0.8234 & 0.7890 & 0.7631 \\ \hline
		COMB & 0.8113 & 0.1150 & 0.8414 & 0.8115 & 0.7971 & 0.7953 & 0.8696 & 0.8236 & 0.7891 & 0.7631 \\ \hline
		ROBUST-ITER & 0.7294 & 0.1330 & 0.7554 & 0.7207 & 0.7190 & 0.7226 & 0.8005 & 0.7445 & 0.7026 & 0.6703 \\ \hline
		ROBUST-COMP & 0.7238 & 0.1348 & 0.7518 & 0.7179 & 0.7126 & 0.7130 & 0.7967 & 0.7381 & 0.6966 & 0.6639 \\ \hline
	\end{tabular}
\end{table}

The number of classes has a significant impact on the classification accuracy attained by different approaches. Table~\ref{tab:accuracy_all} confirms that the accuracies decrease for a greater number of classes. For example, for UTADISMP2 -- the accuracy ranges between $0.8108$ for $p=2$ and $0.7553$ for $p=5$. It is intuitive because, with more classes, the sorting problem becomes more challenging, the sub-spaces of feasible models corresponding to different classes are more constrained, and the class thresholds become closer to each other. As a result, the comprehensive values of non-reference alternatives have lower chances to fit in the value range corresponding to their expected class. 
The greatest decrease in performance is observed between problems with $2$ and $3$ classes (from $2.42\%$ for UTADISMP1 to $6.81\%$ for MIN-SVF). However, with the increasing number of classes, these differences become lesser, and when comparing the results for $4$ and $5$ for some procedures -- they are negligible.

Compared to other methods, a marginal decrease of accuracy with an increasing $p$ is an additional advantage of ACUTADIS. This procedure proves to be more robust to modifying $p$, increasing its relative advantage over the remaining methods when more classes are considered. In the same spirit, the underperformance of MIN-SVF and MAX-SVF is more evident for instances involving more classes. In the conducted experiments, the increase of $p$ implies a greater number of reference alternatives are considered. This directly influences the objective functions of MIN-SVF and MAX-SVF, optimizing the sum of comprehensive values for all alternatives. 

The number of criteria and characteristic points affect the accuracies similarly to the number of classes. With the increase in $m$ and $\gamma_j$, the performance of all procedures deteriorates (see Tables~\ref{tab:accuracy_all} and~\ref{tab:accuracy_characteristic_points}). For example, for CHEBYSHEV, an average accuracy ranges between $0.8623$ and $0.7650$ for $3$ and $9$ criteria, respectively, and between $0.8526$ and $0.7790$ for $2$ and $6$ characteristic points. Again, this is intuitive because, with more criteria and characteristic points, the space of feasible models becomes greater, and MVFs become more flexible. 

The average differences between accuracies for problems with three and nine criteria range from $8.95$ to $17.50\%$ (see Table~\ref{tab:accuracy_all}). Hence, they are more substantial than between the extreme numbers of classes (e.g., for UTADISMP1 and UTADIS-JLS -- even four times greater). The sole exception in this regard is UTADISMP3. The decrease in accuracy is visible between all subsequent numbers of criteria. It is on average $5.1\%$ between $3$ and $5$ criteria, $3.67\%$ between $5$ and $7$ criteria, and $2.88\%$ between $7$ and $9$ criteria. 
As for the number of characteristic points (see Table \ref{tab:accuracy_characteristic_points}), there is a clear difference in the accuracy of methods between linear and piecewise-linear MVFs. The scores attained for MVFs with $2$ and $6$ characteristic points differ from $4.53\%$ for UTADISMP2 up to $14.73\%$ for UTADIS-JLS.

Noteworthy, UTADIS-JLS achieved relatively high results (85.02\% compared to 86.96\% accuracy achieved by the best method -- ACUTADIS) when using linear value functions. However, when employing six characteristic points, the difference between these two methods increased to over $10\%$. Such a difference is associated with optimizing values assigned to the last characteristic points for each MVF. For the linear functions, this contributes to controlling their entire shapes and selecting more central value functions. In turn, with greater $\gamma_j$, the marginal values of intermediate characteristic points are not directly affected by the optimized model. For MSCVF, the differences accuracies attained for MVFs with 4 and 6 points are negligible. This is due to the characteristic of the method, which -- regardless of the number of points -- tries to linearize the marginal functions as much as possible.

\begin{table}[h]
	\caption{Average classification accuracy for different numbers of characteristic points and reference alternatives per class.}\label{tab:accuracy_characteristic_points}\centering\footnotesize
	\begin{tabular} {|l||l|l|l||l|l|l|l|}
		\hline
		& \multicolumn{3}{c||}{Number of ch. points}  & \multicolumn{4}{c|}{Number of reference assignments}\\
		\hline
		\textbf{Procedure} & \textbf{2} & \textbf{4} & \textbf{6}  & \textbf{3} & \textbf{5} & \textbf{7} & \textbf{10} \\ \hline
		UTADISMP1 & 0.8446 & 0.7739 & 0.7507 & 0.7069 & 0.7772 & 0.8200 & 0.8547 \\ \hline
		UTADISMP2 & 0.7993 & 0.7735 & 0.7540 & 0.6948 & 0.7631 & 0.8042 & 0.8405 \\ \hline
		UTADISMP3 & 0.7869 & 0.7363 & 0.7217 & 0.6695 & 0.7340 & 0.7753 & 0.8145 \\ \hline
		UTADIS-JLS & 0.8502 & 0.7579 & 0.7029 & 0.6695 & 0.7559 & 0.8061 & 0.8498 \\ \hline
		CHEBYSHEV & 0.8526 & 0.7980 & 0.7790 & 0.7423 & 0.7987 & 0.8341 & 0.8644 \\ \hline
		MAX-SVF & 0.7330 & 0.6526 & 0.6476 & 0.5881 & 0.6592 & 0.7042 & 0.7596 \\ \hline
		MIN-SVF & 0.7262 & 0.6527 & 0.6466 & 0.5858 & 0.6545 & 0.7057 & 0.7547 \\ \hline
		MSCVF &  & 0.7096 & 0.7105 & 0.6113 & 0.6986 & 0.7442 & 0.7861 \\ \hline
		ACUTADIS & 0.8696 & 0.8184 & 0.8059 & 0.7713 & 0.8232 & 0.8520 & 0.8788 \\ \hline
		CENTROID & 0.8673 & 0.7979 & 0.7748 & 0.7459 & 0.8021 & 0.8380 & 0.8674 \\ \hline
		REPDIS & 0.8126 & 0.7376 & 0.7212 & 0.6763 & 0.7437 & 0.7848 & 0.8237 \\ \hline
		CAI & 0.8662 & 0.7962 & 0.7733 & 0.7439 & 0.8007 & 0.8365 & 0.8665 \\ \hline
		APOI & 0.8649 & 0.7958 & 0.7731 & 0.7421 & 0.8003 & 0.8362 & 0.8665 \\ \hline
		COMB & 0.8650 & 0.7960 & 0.7731 & 0.7422 & 0.8004 & 0.8363 & 0.8665 \\ \hline
		ROBUST-ITER & 0.7927 & 0.7098 & 0.6858& 0.6397 & 0.7107 & 0.7588 & 0.8085 \\ \hline
		ROBUST-COMP & 0.7905 & 0.7036 & 0.6774 & 0.6352 & 0.7057 & 0.7543 & 0.8001 \\ \hline
	\end{tabular}
\end{table}

The increase in the number of reference alternatives per class positively affects the classification accuracy (see Table \ref{tab:accuracy_characteristic_points}). For example, for UTADISMP1, the accuracy ranges between $0.7069$ and $0.8547$ for, respectively, $R=3$ and $10$. A greater number of assignment examples makes the knowledge available to the methods more complete, offering additional arguments on the DM's sorting policy. From a mathematical viewpoint, additional indirect statements constrain the space of feasible models, leaving lesser freedom to the procedures for selecting a representative model. 

With limited preference information (see $R=3$), ACUTADIS has a clear advantage over the remaining methods (over $2.5\%$ over CENTROID). In general, the margin between the stochastic- (CENTROID, CAI, APOI, COMB) or centralization-based (ACUTADIS,  CENTROID, CHEBYSHEV) and the remaining approaches is greater with more sparse DM's preferences. For example, for $R=3$ -- the difference in accuracies of APOI and UTADISMP1 is $3.52\%$, whereas for $R=10$ -- it drops to $1.18\%$. This emphasizes the usefulness of the best-performing approaches when only a few assignment examples are available. 

\subsubsection{Assignment acceptability}
\label{subsec:assignments_acceptability_results}

\noindent Average assignment acceptabilities over all problem instances are provided in Table \ref{tab:mcai_all}. The difference between the best and worst-performing procedures is enormous (almost $0.25$). The procedures exploiting stochastic acceptabilities attained the highest absolute MCAIs. In particular, the CAI procedure aims at emphasizing the most frequent assignments when selecting a representative model. It is very successful in attaining this target with absolute MCAI equal to $0.8979$ and its relative counterpart being close to zero. This means that the CAI method identifies a model that classifies all alternatives according to the \emph{robust assignment rule}~\cite{Doumpos2014}, i.e., it assigns each alternative to a class associated with the highest CAI. Since for some problem instances, there was no model optimizing such an objective in a~perfect way (in the experiment -- this happened for 4 out of 19 200 instances), the relative MCAI for this method does not reach zero. The APOI and COMB methods are only marginally worse in this regard (absolute MCAI equal to 0.8975). This means that considering stochastic acceptabilities for the assignment-based pairwise preference relations led to different assignments for very few problem instances. This confirms that the two perspectives are highly consistent in guiding the methods to the most robust assignments. 

Another group of methods that perform well in terms of assigning alternatives to their most frequent classes in the set of all feasible models is composed of CENTROID ($0.8968$), CHEBYSHEV ($0.8620$), and ACUTADIS ($0.8449$). Note that CAI, APOI, and COMB have a competitive advantage over these methods in considering the assignments of all alternatives, including non-reference ones, already at the stage of identifying a representative model. When such an approach is too costly in terms of required computational effort, one can opt for methods selecting a central model that exploit only the information provided by the DM. Interestingly, unlike for the classification accuracy, ACUTADIS performs slightly worse than procedures selecting an average model or the Chebyshev center. 

\begin{table}[h]
	\caption{Mean values and standard deviations of assignment acceptabilities for all considered problem settings.}\label{tab:mcai_all}\centering\footnotesize
	\begin{tabular}{|l||l|l||l|l|}
		\hline
		\textbf{} & \multicolumn{2}{c||}{\textbf{Absolute}} & \multicolumn{2}{c|}{\textbf{Relative}} \\ \hline
		\textbf{Procedure} & \textbf{mean} & \textbf{std} & \textbf{mean} & \textbf{std} \\ \hline
		UTADISMP1 & 0.8034 & 0.1058 & -0.1060 & 0.1010  \\ \hline
		UTADISMP2 & 0.7902 & 0.1046 & -0.1208 & 0.0998 \\ \hline
		UTADISMP3 & 0.7814 & 0.1005 & -0.1312 & 0.0887 \\ \hline
		UTADIS-JLS & 0.7766 & 0.1191 & -0.1357 & 0.1193 \\ \hline
		CHEBYSHEV & 0.8620 & 0.0669 & -0.0407 & 0.0396 \\ \hline
		MAX-SVF & 0.6771 & 0.1444 & -0.2480 & 0.1451 \\ \hline
		MIN-SVF & 0.6763 & 0.1430 & -0.2489 & 0.1435 \\ \hline
		MSCVF & 0.7161 & 0.1209 & -0.2095 & 0.1202 \\ \hline
		ACUTADIS & 0.8449 & 0.0760 & -0.0594 & 0.0622 \\ \hline
		CENTROID & 0.8968 & 0.0491 & -0.0012 & 0.0025 \\ \hline
		REPDIS & 0.7980 & 0.0902 & -0.1123 & 0.0782 \\ \hline
		CAI & 0.8979 & 0.0485 & -1.4E-07 & 1.4E-05 \\ \hline
		APOI & 0.8975 & 0.0491 & -0.0005 & 0.0020 \\ \hline
		COMB & 0.8975 & 0.0490 & -0.0005 & 0.0018 \\ \hline
		ROBUST-ITER & 0.7400 & 0.1251 & -0.1773 & 0.1228 \\ \hline
		ROBUST-COMP & 0.7272 & 0.1261 & -0.1913 & 0.1256 \\ \hline
	\end{tabular}
\end{table}

The worst performers in terms of assignment acceptability are the same as for the classification accuracy. The least robustness of recommended assignments is observed for MIN-SVF (0.6771), MAX-SVF (0.6763), and MSCVF (0.7161). It is understandable given the objective functions optimized by these approaches that have nothing in common with alternatives' assignments or robustness of results. In this regard, surprisingly low MCAIs are attained by ROBUST-ITER (0.7400) and ROBUST-COMP (0.7272). These procedures build on the outcomes of robustness analysis. However, they focus on the necessary and possible relations derived from mathematical programming. This proves that such extreme, robust outcomes are often too scarce to provide valuable insights and guide the procedures to select a model that would be representative in terms of robustness preoccupation. Since the space between the necessary and the possible may be quite large, the use of stochastic acceptabilities computed with the Monte Carlo simulation and filling this gap turns out more beneficial for most problem instances. 

In Tables~\ref{tab:mcai_classes}--\ref{tab:mcai_ref_alternatives}, we provide the average assignment acceptabilities for different values of particular dimensions. In general, the robustness of recommended assignments increases with fewer classes, criteria, and characteristic points and a greater number of reference alternatives per class. Hence, these trends are the same as for the classification accuracy. They can be attributed to the same reasons. Less flexible models and greater information load lead to more constrained space of feasible models and more robust sorting results. For example, for UTADISMP2 -- the difference between extreme values for each dimension are as follows: for $p \in \{2, 5\}$ -- 0.0359, for $m \in \{3, 9\}$ -- 0.0908, for $\gamma_j \in \{2, 6\}$ -- 0.0221, and for $R \in \{3, 10\}$ -- 0.1539. This indicates that the number of reference alternatives per class has the greatest impact on the robustness of recommended assignments. In contrast, the influence of the number of characteristic points is the least.

When it comes to absolute MCAIs attained for different numbers of classes (see Table \ref{tab:mcai_classes}), the greatest differences are observed for problems with $2$ and $3$ classes. The deviation from the general trend is noted for some procedures when comparing the results for problems with $4$ and $5$ classes. As far as various procedures are concerned, the performance of CAI, APOI, COMB, and CENTROID is the most stable (e.g., for the last approach, absolute MCAI is $0.9099$ for $p=2$ and $0.8931$ for $p=5$). All these methods share the component of performing the stochastic acceptability analysis. In turn, the disadvantage of the worst performers (MAX-SVF, MIN-SVF, and MSCVF) is more evident with increased problem complexity. This is captured by the relative MCAIs, where their loss to the most robust possible assignments increases from about $0.18$ to $0.23-0.29$. For the remaining approaches, these relative values are more stable, and for some of them (see, e.g., UTADISMP1 and REPDIS), they tend to be smaller when moving from three to five classes. 

\begin{table}[h]
	\caption{Average assignment acceptability for different numbers of classes.}\label{tab:mcai_classes}\centering\footnotesize
	\begin{tabular}{|l||l|l|l|l||l|l|l|l|}
		\hline
		\textbf{} & \multicolumn{4}{c||}{\textbf{Absolute}} & \multicolumn{4}{c|}{\textbf{Relative}} \\ \hline
		\textbf{Procedure} & \textbf{2} & \textbf{3} & \textbf{4} & \textbf{5} & \textbf{2} & \textbf{3} & \textbf{4} & \textbf{5} \\ \hline
		UTADISMP1 & 0.8155 & 0.7949 & 0.7982 & 0.8048 & -0.1048 & -0.1120 & -0.1060 & -0.1013 \\ \hline
		UTADISMP2 & 0.8158 & 0.7857 & 0.7793 & 0.7799 & -0.1039 & -0.1223 & -0.1274 & -0.1295 \\ \hline
		UTADISMP3 & 0.8421 & 0.7840 & 0.7549 & 0.7446 & -0.0757 & -0.1247 & -0.1551 & -0.1695 \\ \hline
		UTADIS-JLS & 0.8050 & 0.7675 & 0.7642 & 0.7698 & -0.1154 & -0.1424 & -0.1442 & -0.1410 \\ \hline
		CHEBYSHEV & 0.8830 & 0.8592 & 0.8525 & 0.8535 & -0.0310 & -0.0403 & -0.0449 & -0.0464 \\ \hline
		MAX-SVF & 0.7456 & 0.6748 & 0.6467 & 0.6415 & -0.1819 & -0.2473 & -0.2771 & -0.2856 \\ \hline
		MIN-SVF & 0.7453 & 0.6733 & 0.6479 & 0.6388 & -0.1824 & -0.2489 & -0.2757 & -0.2886 \\ \hline
		MSCVF & 0.7645 & 0.7175 & 0.6993 & 0.6831 & -0.1728 & -0.2058 & -0.2210 & -0.2385 \\ \hline
		ACUTADIS & 0.8744 & 0.8410 & 0.8319 & 0.8321 & -0.0399 & -0.0602 & -0.0676 & -0.0702 \\ \hline
		CENTROID & 0.9099 & 0.8936 & 0.8906 & 0.8931 & -0.0009 & -0.0013 & -0.0014 & -0.0013 \\ \hline
		REPDIS & 0.8132 & 0.7910 & 0.7918 & 0.7959 & -0.1079 & -0.1168 & -0.1131 & -0.1112 \\ \hline
		CAI & 0.9107 & 0.8947 & 0.8918 & 0.8943 & 0 & -1.9E-08 & -4.1E-07 & -1.5E-07 \\ \hline
		APOI & 0.9104 & 0.8942 & 0.8913 & 0.8939 & -0.0003 & -0.0006 & -0.0006 & -0.0005 \\ \hline
		COMB & 0.9105 & 0.8943 & 0.8914 & 0.8939 & -0.0003 & -0.0006 & -0.0005 & -0.0004 \\ \hline
		ROBUST-ITER & 0.7597 & 0.7298 & 0.7313 & 0.7390 & -0.1661 & -0.1854 & -0.1818 & -0.1757 \\ \hline
		ROBUST-COMP & 0.7563 & 0.7205 & 0.7167 & 0.7153 & -0.1698 & -0.1954 & -0.1978 & -0.2021 \\ \hline
	\end{tabular}
\end{table}

When increasing the number of criteria, the trends for absolute and relative MCAIs are more consistent (see Table~\ref{tab:mcai_criteria}). For all procedures, the robustness of recommended assignments decreases in terms of absolute values and their distances from the best possible solution. Also, with more criteria, the performance differences become sharper in both absolute and relative terms. For example, when considering the extreme numbers of criteria $m \in \{3, 9\}$, the absolute loss of MIN-SVF to CAI increases from $0.1379$ to $0.2893$, and the respective relative loss grows from $0.1542$ to $0.3249$.

\begin{table}[h]
	\caption{Average assignment acceptability for different numbers of criteria.}\label{tab:mcai_criteria}\centering\footnotesize
	\begin{tabular}{|l||l|l|l|l||l|l|l|l|}
		\hline
		& \multicolumn{4}{c||}{\textbf{Absolute}} & \multicolumn{4}{c|}{\textbf{Relative}} \\ \hline
		\textbf{Procedure} & \textbf{3} & \textbf{5} & \textbf{7} & \textbf{9} & \textbf{3} & \textbf{5} & \textbf{7} & \textbf{9} \\ \hline
		UTADISMP1 & 0.8593 & 0.8163 & 0.7830 & 0.7549 & -0.0540 & -0.0909 & -0.1248 & -0.1545 \\ \hline
		UTADISMP2 & 0.8379 & 0.8030 & 0.7727 & 0.7471 & -0.0774 & -0.1058 & -0.1365 & -0.1634 \\ \hline
		UTADISMP3 & 0.8163 & 0.7867 & 0.7686 & 0.7541 & -0.1018 & -0.1243 & -0.1419 & -0.1570 \\ \hline
		UTADIS-JLS & 0.8231 & 0.7886 & 0.7607 & 0.7340 & -0.0943 & -0.1215 & -0.1494 & -0.1776 \\ \hline
		CHEBYSHEV & 0.8821 & 0.8635 & 0.8549 & 0.8476 & -0.0283 & -0.0380 & -0.0446 & -0.0517 \\ \hline
		MAX-SVF & 0.7722 & 0.6933 & 0.6413 & 0.6017 & -0.1512 & -0.2291 & -0.2843 & -0.3272 \\ \hline
		MIN-SVF & 0.7694 & 0.6907 & 0.6413 & 0.6038 & -0.1542 & -0.2319 & -0.2846 & -0.3249 \\ \hline
		MSCVF & 0.7896 & 0.7280 & 0.6876 & 0.6592 & -0.1272 & -0.1932 & -0.2417 & -0.2761 \\ \hline
		ACUTADIS & 0.8794 & 0.8509 & 0.8323 & 0.8169 & -0.0313 & -0.0518 & -0.0694 & -0.0853 \\ \hline
		CENTROID & 0.9065 & 0.8960 & 0.8930 & 0.8918 & -0.0009 & -0.0012 & -0.0013 & -0.0015 \\ \hline
		REPDIS & 0.8301 & 0.8019 & 0.7853 & 0.7745 & -0.0863 & -0.1069 & -0.1225 & -0.1334 \\ \hline
		CAI & 0.9073 & 0.8970 & 0.8941 & 0.8931 & 0 & -1.7E-07 & 0 & -4.1E-07 \\ \hline
		APOI & 0.9069 & 0.8966 & 0.8937 & 0.8927 & -0.0005 & -0.0005 & -0.0005 & -0.0006 \\ \hline
		COMB & 0.9069 & 0.8967 & 0.8937 & 0.8927 & -0.0004 & -0.0004 & -0.0005 & -0.0005 \\ \hline
		ROBUST-ITER & 0.8081 & 0.7561 & 0.7134 & 0.6823 & -0.1114 & -0.1586 & -0.2031 & -0.2359 \\ \hline
		ROBUST-COMP & 0.7957 & 0.7405 & 0.7009 & 0.6717 & -0.1250 & -0.1758 & -0.2167 & -0.2477 \\ \hline
	\end{tabular}
\end{table}

The general trend of decreasing absolute MCAIs with a greater number of characteristic points is visible in Table~\ref{tab:mcai_characteristic_points}. However, it is not valid for all procedures. For the best-performing methods, including CAI, APOI, COMB, CHEBYSHEV, and CENTROID, it is inverse. For example, the absolute MCAIs for CAI are $0.8849$, $0.8942$, and $0.9145$ for $\gamma_j = 2, 4, 6$. For the approaches exploiting the stochastic acceptabilities, one may interpret that more flexible MVFs offer greater chances for better fitting the models to reflect the CAIs and APOIs. Even though for the procedures identifying the Chebyshev and analytic centers the absolute MCAIs increased when moving from linear to piecewise linear MVFs, their relative counterparts marginally deteriorated.

\begin{table}[h]
	\caption{Average assignment acceptability for different numbers of characteristic points.}\label{tab:mcai_characteristic_points}\centering\footnotesize
	\begin{tabular}{|l||l|l|l||l|l|l|}
		\hline
		\textbf{} & \multicolumn{3}{c||}{\textbf{Absolute}} & \multicolumn{3}{c|}{\textbf{Relative}} \\ \hline
		\textbf{Procedure} & \textbf{2} & \textbf{4} & \textbf{6} & \textbf{2} & \textbf{4} & \textbf{6} \\ \hline
		UTADISMP1 & 0.8484 & 0.7966 & 0.7650 & -0.0428 & -0.1108 & -0.1645 \\ \hline
		UTADISMP2 & 0.7949 & 0.8029 & 0.7728 & -0.1032 & -0.1034 & -0.1557 \\ \hline
		UTADISMP3 & 0.7828 & 0.7711 & 0.7904 & -0.1175 & -0.1393 & -0.1369 \\ \hline
		UTADIS-JLS & 0.8460 & 0.7667 & 0.7171 & -0.0457 & -0.1444 & -0.2171 \\ \hline
		CHEBYSHEV & 0.8568 & 0.8540 & 0.8753 & -0.0328 & -0.0457 & -0.0434 \\ \hline
		MAX-SVF & 0.7310 & 0.6518 & 0.6486 & -0.1781 & -0.2736 & -0.2922 \\ \hline
		MIN-SVF & 0.7270 & 0.6537 & 0.6482 & -0.1826 & -0.2716 & -0.2925 \\ \hline
		MSCVF &  & 0.7160 & 0.7162 &  & -0.2013 & -0.2178 \\ \hline
		ACUTADIS & 0.8646 & 0.8326 & 0.8374 & -0.0237 & -0.0698 & -0.0848 \\ \hline
		CENTROID & 0.8842 & 0.8931 & 0.9132 & -0.0009 & -0.0014 & -0.0014 \\ \hline
		REPDIS & 0.8199 & 0.7847 & 0.7893 & -0.0753 & -0.1235 & -0.1379 \\ \hline
		CAI & 0.8849 & 0.8942 & 0.9145 & -3.2E-07 & -1.1E-07 & 0 \\ \hline
		APOI & 0.8844 & 0.8938 & 0.9142 & -0.0007 & -0.0005 & -0.0003 \\ \hline
		COMB & 0.8844 & 0.8939 & 0.9142 & -0.0006 & -0.0005 & -0.0003 \\ \hline
		ROBUST-ITER & 0.7972 & 0.7262 & 0.6965 & -0.1020 & -0.1902 & -0.2396 \\ \hline
		ROBUST-COMP & 0.7883 & 0.7105 & 0.6828 & -0.1119 & -0.2075 & -0.2545 \\ \hline
	\end{tabular}
\end{table}

With a more significant number of reference alternatives per class, the trends of increasing absolute MCAI and decreasing loss to the most robust assignment are unanimously confirmed for all procedures (see Table~\ref{tab:mcai_ref_alternatives}. For example, for CHEBYSHEV, its absolute assignment acceptability increases from $0.8105$ to $0.9044$ when moving from $R=3$ to $10$, and its relative loss decreases from $0.0572$ to $0.0277$. With additional preference information, the entropy of class acceptability indices gets lower, and hence the feasible models become more similar in terms of the suggested sorting recommendations~\cite{Kadzinski21ALSORT}. Consequently, irrespective of the applied procedure, its chances for selecting a model whose assignments are highly robust get higher. 

\begin{table}[h]
	\caption{Average assignment acceptability for different numbers of reference alternatives per class.}\label{tab:mcai_ref_alternatives}\centering\footnotesize
	\begin{tabular}{|l||l|l|l|l||l|l|l|l|}
		\hline
		\textbf{} & \multicolumn{4}{c||}{\textbf{Absolute}} & \multicolumn{4}{c|}{\textbf{Relative}} \\ \hline
		\textbf{Procedure} & \textbf{3} & \textbf{5} & \textbf{7} & \textbf{10} & \textbf{3} & \textbf{5} & \textbf{7} & \textbf{10} \\ \hline
		UTADISMP1 & 0.7170 & 0.7914 & 0.8344 & 0.8707 & -0.1641 & -0.1119 & -0.0843 & -0.0639 \\ \hline
		UTADISMP2 & 0.7040 & 0.7776 & 0.8211 & 0.8579 & -0.1794 & -0.1274 & -0.0987 & -0.0776 \\ \hline
		UTADISMP3 & 0.7110 & 0.7695 & 0.8052 & 0.8399 & -0.1735 & -0.1375 & -0.1168 & -0.0971 \\ \hline
		UTADIS-JLS & 0.6785 & 0.7624 & 0.8102 & 0.8553 & -0.2085 & -0.1439 & -0.1104 & -0.0802 \\ \hline
		CHEBYSHEV & 0.8105 & 0.8535 & 0.8797 & 0.9044 & -0.0572 & -0.0430 & -0.0348 & -0.0277 \\ \hline
		MAX-SVF & 0.5868 & 0.6579 & 0.7052 & 0.7586 & -0.3179 & -0.2626 & -0.2266 & -0.1848 \\ \hline
		MIN-SVF & 0.5865 & 0.6556 & 0.7067 & 0.7565 & -0.3183 & -0.2653 & -0.2250 & -0.1870 \\ \hline
		MSCVF & 0.6232 & 0.7041 & 0.7479 & 0.7893 & -0.2865 & -0.2164 & -0.1828 & -0.1524 \\ \hline
		ACUTADIS & 0.7939 & 0.8367 & 0.8618 & 0.8870 & -0.0756 & -0.0614 & -0.0543 & -0.0464 \\ \hline
		CENTROID & 0.8575 & 0.8903 & 0.9102 & 0.9293 & -0.0019 & -0.0013 & -0.0010 & -0.0007 \\ \hline
		REPDIS & 0.7288 & 0.7875 & 0.8213 & 0.8542 & -0.1517 & -0.1167 & -0.0989 & -0.0818 \\ \hline
		CAI & 0.8591 & 0.8914 & 0.9111 & 0.9299 & -5.3E-07 & -5.3E-08 & 0 & 0 \\ \hline
		APOI & 0.8580 & 0.8911 & 0.9110 & 0.9299 & -0.0013 & -0.0004 & -0.0002 & -0.0001 \\ \hline
		COMB & 0.8581 & 0.8911 & 0.9110 & 0.9299 & -0.0012 & -0.0004 & -0.0002 & -0.0001 \\ \hline
		ROBUST-ITER & 0.6480 & 0.7213 & 0.7704 & 0.8202 & -0.2452 & -0.1908 & -0.1547 & -0.1183 \\ \hline
		ROBUST-COMP & 0.6378 & 0.7087 & 0.7581 & 0.8042 & -0.2570 & -0.2048 & -0.1680 & -0.1354 \\ \hline
	\end{tabular}
\end{table}

\subsubsection{Differences between marginal and comprehensive values and class thresholds}
\label{subsec:criteria_tradeoffs_results}
\noindent In this section, we will discus the results for the remaining three measures jointly, because the underlying ranking of methods are similar to a large extent. This is understandable, because all measures concern the similarity between the models derived with different approaches and the reference model, though referring to its various components. We present the average differences between marginal and comprehensive values and class thresholds in Table~\ref{tab:tradeoffs_all}. In additional, for the marginal values, we report the difference to an average solution obtained with CENTROID. 

The most significant similarity to the reference model is observed for the outcomes of procedures derived with ACUTADIS, CENTROID, and CHEBYSHEV. Such an order is confirmed for the three measures. For example, for ACUTADIS, the distance in terms of marginal values is $0.0417$, for comprehensive values -- it is $0.0502$, and for class thresholds -- $0.0409$. For the procedures identifying an average solution and the Chebyshev center, values for these measures are only slightly higher. The distances of the function returned by ACUTADIS and CHEBYSHEV from the centroid solution are very low, suggesting that the three procedures return similar models. For UTADIS-JLS, which implements an analogous selection rule to CENTROID, such a distance is higher, which is also reflected in more substantial differences from the DM's reference model.

Favorable results in terms of differences between marginal and comprehensive values are attained with REPDIS. This procedure builds its model on the Assignment-based Pairwise Winning Indices. However, when considering the class thresholds, these differences are higher. It is intuitive because REPDIS does not optimize the threshold values, focusing only on selecting a representative value function. Still, REPDIS proves better in terms of the three measures than the remaining methods exploiting the stochastic acceptabilities. For example, for CAI, the distance from the reference model in terms of marginal values is $0.0642$, for comprehensive values -- it is $0.0916$, and for class thresholds -- $0.0818$, being $1.5-2$ times higher than for ACUTADIS. This confirms that aiming to reproduce the most common results attained in the set of all compatible sorting models does not allow perfectly replicating a~single reference model.

UTADISMP1 and ROBUST-ITER achieve the intermediate results. These procedures attain exactly the same values in terms of distances built on marginal and comprehensive values. This is because they aim at identifying the most discriminant models. While UTADISMP1 exploits only the DM's preference information, ROBUST-ITER refers to the necessary assignment-based preference relation. However, this relation is heavily influenced by the DM's assignment examples in the sense that all reference alternatives from the more preferred classes are necessarily preferred to the alternatives assigned to the least preferred classes. The differences between these methods can be observed for the measure values related to class thresholds. This is because UTADISMP1 directly optimizes their values, while ROBUST-ITER is focused only on the parameters of the AVF. Furthermore, UTADISMP2 constructed models that are, on average, slightly more similar to the reference one than UTADISMP1, whereas the similarity results for ROBUST-COMP are marginally worse than for its iterative counterpart.

Finally, the furthest models from the reference one were obtained with MAX-SVF and MIN-SVF. This is confirmed by the distances concerning marginal and comprehensive values as well as class thresholds (e.g., for MAX-SVF -- they are $0.1373$, $0.2391$, and $0.2671$). Such high values do not only indicate a significant dissimilarity of the benevolent and aggressive models from the simulated DM's model, but they also prove a large variability of models that are compatible with the DM's incomplete preference information.

\begin{table}[h]
	\caption{Average value and standard deviation of differences between marginal values, comprehensive values, and class thresholds from the reference model (in case of marginal values, also the differences from the centroid model).
}\label{tab:tradeoffs_all}\centering\footnotesize
	\begin{tabular} {|l||l|l||l|l|| l|l||  l|l|}
		\hline
			& \multicolumn{4}{c||}{\textbf{Marginal values}} & \multicolumn{2}{c||}{\textbf{Comprehensive}}  & \multicolumn{2}{c|}{\textbf{Class}}  \\
		& \multicolumn{2}{c}{\textbf{Reference}} & \multicolumn{2}{c||}{\textbf{Centroid}} & \multicolumn{2}{c||}{\textbf{values}}  & \multicolumn{2}{c|}{\textbf{thresholds}}  \\ \hline
		\textbf{Procedure} & \textbf{mean} & \textbf{std} & \textbf{mean} & \textbf{std} & \textbf{mean} & \textbf{std} & \textbf{mean} & \textbf{std} \\ \hline
		UTADISMP1 & 0.0582 & 0.0439 & 0.0493 & 0.0412 & 0.0811 & 0.0709  & 0.0594 & 0.0602 \\ \hline
		UTADISMP2 & 0.0606 & 0.0412 & 0.0516 & 0.0396& 0.0803 & 0.0679& 0.0588 & 0.0572 \\ \hline
		UTADISMP3 & 0.0607 & 0.0362 & 0.0424 & 0.0269 & 0.0655 & 0.0367 & 0.0526 & 0.0440 \\ \hline
		UTADIS-JLS & 0.0558 & 0.0403 & 0.0426 & 0.0330 & 0.0997 & 0.0887 & 0.1004 & 0.1024 \\ \hline
		CHEBYSHEV & 0.0459 & 0.0313 & 0.0207 & 0.0174 & 0.0553 & 0.0375 & 0.0461 & 0.0436 \\ \hline
		MAX-SVF & 0.1373 & 0.0958 & 0.1381 & 0.0957 & 0.2391 & 0.1569 & 0.2671 & 0.1958 \\ \hline
		MIN-SVF & 0.1003 & 0.0594 & 0.0989 & 0.0567 & 0.2391 & 0.1562 & 0.2682 & 0.1940 \\ \hline
		MSCVF & 0.0628 & 0.0384 & 0.0547 & 0.0312 & 0.0870 & 0.0421 & 0.0758 & 0.0551 \\ \hline
		ACUTADIS & 0.0417 & 0.0286 & 0.0239 & 0.0158 & 0.0502 & 0.0337 & 0.0409 & 0.0395 \\ \hline
		CENTROID & 0.0445 & 0.0291 & 0 & 0 & 0.0545 & 0.0380 & 0.0461 & 0.0449 \\ \hline
		REPDIS & 0.0492 & 0.0357 & 0.0256 & 0.0262 & 0.0622 & 0.0519 & 0.0684 & 0.0680 \\ \hline
		CAI & 0.0642 & 0.0375 & 0.0463 & 0.0256 & 0.0916 & 0.0801 & 0.0818 & 0.0877 \\ \hline
		APOI & 0.0644 & 0.0377 & 0.0465 & 0.0256 & 0.0908 & 0.0790 & 0.0812 & 0.0868 \\ \hline
		COMB & 0.0644 & 0.0377 & 0.0465 & 0.0256 & 0.0908 & 0.0790 & 0.0812 & 0.0868 \\ \hline
		ROBUST-ITER & 0.0582 & 0.0439 & 0.0493 & 0.0412 & 0.0811 & 0.0709 & 0.0958 & 0.1023 \\ \hline
		ROBUST-COMP & 0.0614 & 0.0454 & 0.0541 & 0.0425  & 0.0853 & 0.0741 & 0.0978 & 0.1035 \\ \hline
		
	\end{tabular}
\end{table}

The differences between the reference and resulting models obtained for different values of each problem dimension ($p$, $m$, $\gamma_j$, and $R$) are discussed in the eAppendix (supplementary material available online).

\section{Summary and future research}
\label{sec:summary}
\noindent We considered preference disaggregation in the context of multiple criteria sorting. We assumed the classification is driven by an additive value function and thresholds separating the categories. The parameters of such a model are inferred from the Decision Maker's assignment examples. The use of such indirect and incomplete preference information leads to infinitely many compatible sorting models, potentially implying different assignments for the non-reference alternatives. Given the multiplicity of feasible models, selecting a single, representative one can be conducted in different ways. 

We reviewed several procedures for such a selection. They aim at identifying the most discriminant, average, central, benevolent, aggressive, parsimonious, or robust model. These ideas differ in terms of the exploited information and aspects to be emphasized that translate into the relevant constraints and an objective function. In this paper, we introduced five novel procedures. Two of them are relatively simple, striving to obtain the most discriminant value function in terms of the shape of marginal functions or an aggressive function that puts all alternatives jointly in the worst possible light. However, our core contribution consists of proposing three novel procedures that aim at assigning the alternatives according to the robust assignment rule. For this purpose, they exploit class acceptability indices and/or assignment-based pairwise acceptabilities and maximize the support given to the resulting assignments by all feasible sorting models. The use of all approaches, including the existing and novel ones, was illustrated in a study concerning the green performance assessment of European cities.

In the extensive experimental study, we compare the performance of all procedures on problem instances with different complexities. The results were quantified in terms of five measures. When it comes to reproducing the assignments generated by a simulated Decision Maker's model and the parameters of this model, involving marginal and comprehensive values as well as class thresholds, the best performers are the same. They include the procedures that determine a central sorting model with the proviso that it can be an analytic center, the Chebyshev center, or an average determined based on a large sample of compatible models. Favorable results in terms of classification accuracy were also attained with the procedures exploiting the stochastic acceptabilities, whereas the most discriminant procedures better approximated the unknown model parameters. 

The novel stochastic approaches proved to be the best in emphasizing the robustness of results in a univocal recommendation. This is, however, at the increased computational cost related to conducting robustness analysis for all alternatives and solving a more challenging optimization problem. As proven by the experimental results, the center-oriented procedures also achieved high robustness of results. On the contrary, optimizing comprehensive values of all alternatives, focussing only on the shape of marginal value functions, or exploiting the exact, necessary results did not lead to favorable outcomes given any considered measure. 

The experimental study indicated that the classification accuracy of procedures and assignment acceptability of their recommendation decreased with more classes, criteria, and characteristic points and fewer reference alternatives per class. These outcomes can be justified given a more significant challenge posed by the classification problems with more classes, higher flexibility of a preference model with more criteria and breakpoints, and greater information gain offered by additional assignment examples. The average differences between the reference and delivered models given values of parameters such as marginal values assigned to particular characteristic points, alternatives' comprehensive values, or class thresholds exhibit slightly different trends. They become lower with more classes (also implying more assignment examples) and reference assignments per class and higher with more characteristic points. Regarding the impact of the number of criteria, the observed regularities were unclear and differed from one approach to another. In most cases, the trend of change in values of all measures was non-linear with respect to considered values of different dimensions. Specifically, greater modifications were observed in the lower scale range of different parameters of a decision problem or a sorting model (e.g., when passing from $2$ to $3$ classes, from $3$ to $5$ criteria, from $2$ to $4$ characteristic points, of from $3$ to $5$ reference assignments per class). In turn, the differences in the upper parts of the parameter scales were lesser (e.g., when passing from $4$ to $5$ classes, from $7$ to $9$ criteria, from $4$ to $6$ characteristic points, of from $7$ to $10$ reference assignments per class).

We envisage the following directions for future research. Firstly, in this paper, we focused only on analyzing procedures for selecting a representative sorting model in case of compatibility with the DM's preference information. However, it would be useful to extend the study in terms of both simulating artificial DMs' policies with the models which do not ensure such a~compatibility as well as considering procedures that are specifically oriented toward selecting a representative model in case of inconsistency~\cite{Beuthe01, Zopounidis00}. Secondly, the study can be brought to the ground of multiple criteria ranking. This would require consideration of a variety of procedures that construct a representative value function based on the DM's pairwise comparisons (see, e.g.,~\cite{Beuthe01, Bous10, Jacquet82UTA}). Moreover, a set of ranking-specific measures would need to be elaborated to quantify the experimental results~\cite{Matsatsinis2018}. Third, it would be interesting to design the procedures compromising between deriving central and robust models. This would allow them to score well in classification accuracy, reproducing the unknown DM's model, and support given to their recommendation in the set of all compatible models.

\section*{Acknowledgments}
\noindent Mi{\l}osz Kadzi{\'n}ski and Micha{\l} W\'ojcik acknowledge financial support from the Polish National Science Center under the SONATA BIS project (grant no. DEC-2019/34/E/HS4/00045). 

%\bibliographystyle{apalike}
%\bibliography{biblio}

\end{document}